\documentclass{article}

\usepackage{microtype}
\usepackage{graphicx}
\usepackage{subfigure}
\usepackage{booktabs} 

\usepackage{threeparttable}
\usepackage[utf8]{inputenc} 
\usepackage[T1]{fontenc}    
\usepackage{algorithmic}
\usepackage{color}
\usepackage{xcolor}
\usepackage{multirow}
\usepackage{hyperref}

\newcommand{\methodname}{SAMG}
\newcommand{\bellman}{\mathcal{B}}
\definecolor{myblue}{rgb}{0.87,0.92,0.96}

\usepackage[accepted]{icml2025}

\usepackage{amsmath}
\usepackage{amssymb}
\usepackage{mathtools}
\usepackage{amsthm}

\usepackage[capitalize,noabbrev]{cleveref}

\theoremstyle{plain}
\newtheorem{theorem}{Theorem}[section]

\theoremstyle{definition}

\theoremstyle{remark}

\usepackage[textsize=tiny]{todonotes}

\icmltitlerunning{SAMG: Offline-to-Online Reinforcement Learning via State-Action-Conditional Offline Model Guidance}

\begin{document}

\twocolumn[
\icmltitle{SAMG: Offline-to-Online Reinforcement Learning via\\ State-Action-Conditional Offline Model Guidance}



\icmlsetsymbol{equal}{*}

\begin{icmlauthorlist}
\icmlauthor{Liyu Zhang}{}
\icmlauthor{Haochi Wu}{}
\icmlauthor{Xu Wan}{}
\icmlauthor{Quan Kong}{}
\icmlauthor{Ruilong Deng}{}
\icmlauthor{Mingyang Sun}{}
\end{icmlauthorlist}



\icmlkeywords{Machine Learning, ICML}

\vskip 0.3in
]




\begin{abstract}
Offline-to-online (O2O) reinforcement learning (RL) pre-trains models on offline data and refines policies through online fine-tuning. However, existing O2O RL algorithms typically require maintaining the tedious offline datasets to mitigate the effects of out-of-distribution (OOD) data, which significantly limits their efficiency in exploiting online samples. To address this deficiency, we introduce a new paradigm for O2O RL called \textbf{S}tate-\textbf{A}ction-Conditional Offline \textbf{M}odel \textbf{G}uidance (SAMG). It freezes the pre-trained offline critic to provide compact offline understanding for each state-action sample, thus eliminating the need for retraining on offline data. The frozen offline critic is incorporated with the online target critic weighted by a state-action-adaptive coefficient. This coefficient aims to capture the offline degree of samples at the state-action level, and is updated adaptively during training. In practice, SAMG could be easily integrated with Q-function-based algorithms. Theoretical analysis shows good optimality and lower estimation error. Empirically, SAMG outperforms state-of-the-art O2O RL algorithms on the D4RL benchmark.
\end{abstract}

\section{Introduction}
\label{Introduction}
Offline reinforcement learning~\cite{lowrey2018plan,fujimoto19a,mao2022moore,rafailov2023moto} has gained significant popularity due to its isolation from online environments. It relies exclusively on offline datasets, which can be generated by one or several policies, constructed from historical data, or even generated randomly. This paradigm eliminates the risks and costs associated with online interactions and offers a safe and efficient pathway to pre-train well-behaved RL agents. However, offline RL algorithms exhibit an inherent limitation in that the offline dataset only covers a partial distribution of the state-action space~\cite{prudencio2023survey}. Therefore, standard online RL algorithms fail to resist the cumulative overestimation on samples out of the offline distribution~\cite{nakamoto2023calql}. To this end, most offline RL algorithms limit the decision-making scope of the estimated policy within the offline dataset distribution~\cite{kumar2019stabilizing,yu2021combo}. Accordingly, offline RL algorithms are conservative and are confined in performance by the limited distribution.

To overcome the performance limitation of offline RL algorithms and further improve their performance, it is inspiring to perform an online fine-tuning process with the offline pre-trained model. Similar to the successful paradigm of transfer learning in deep learning~\cite{weiss2016survey,iman2023review}, this paradigm, categorized as offline-to-online (O2O) RL algorithms, is anticipated to enable substantially faster convergence compared to pure online RL. However, the online fine-tuning process inevitably encounters out-of-distribution (OOD) samples laid aside in the offline pre-training process. This leads to another dilemma: the conservative pre-trained model may be misguided toward structural damage and performance deterioration when coming across OOD samples~\cite{nair2020accelerating,kostrikov2022offline}. Therefore, O2O RL algorithms tend to remain unchanged or even sharply decline in the initial stage of the fine-tuning process. Existing O2O RL algorithms conquer this by maintaining access to the offline dataset and retraining on the offline data during online iterations to restore offline information and restrict OOD deterioration.

Specifically, most fine-tuning algorithms directly inherit the offline dataset as online replay buffer and only get access to online data by incrementally replacing offline data with online ones through iterations~\cite{lyu2022mildly,lee2022offline,wen2023towards,wang2024train}. This paradigm is tedious given that the sample size of the offline datasets tends to exceed the order of millions~\cite{fu2020d4rl}. Hence, these algorithms exhibit low inefficiency in leveraging online data. Other algorithms~\cite{nakamoto2023calql,zheng2023adaptive,guo2023simple,liu2024energyguided} maintain an online buffer and an offline one and sample from the two replay buffers with hybrid setting~\cite{song2023hybrid} or priority sampling technique. Though these settings mitigate the inefficiency, they still visit a considerable amount of offline data and have not departed from the burden of offline data. In summary, existing O2O RL algorithms severely compromise the efficiency of utilizing online data to mitigate the negative impact of OOD samples.
\begin{figure}[t]
\centering
\includegraphics[width=0.38\textwidth]{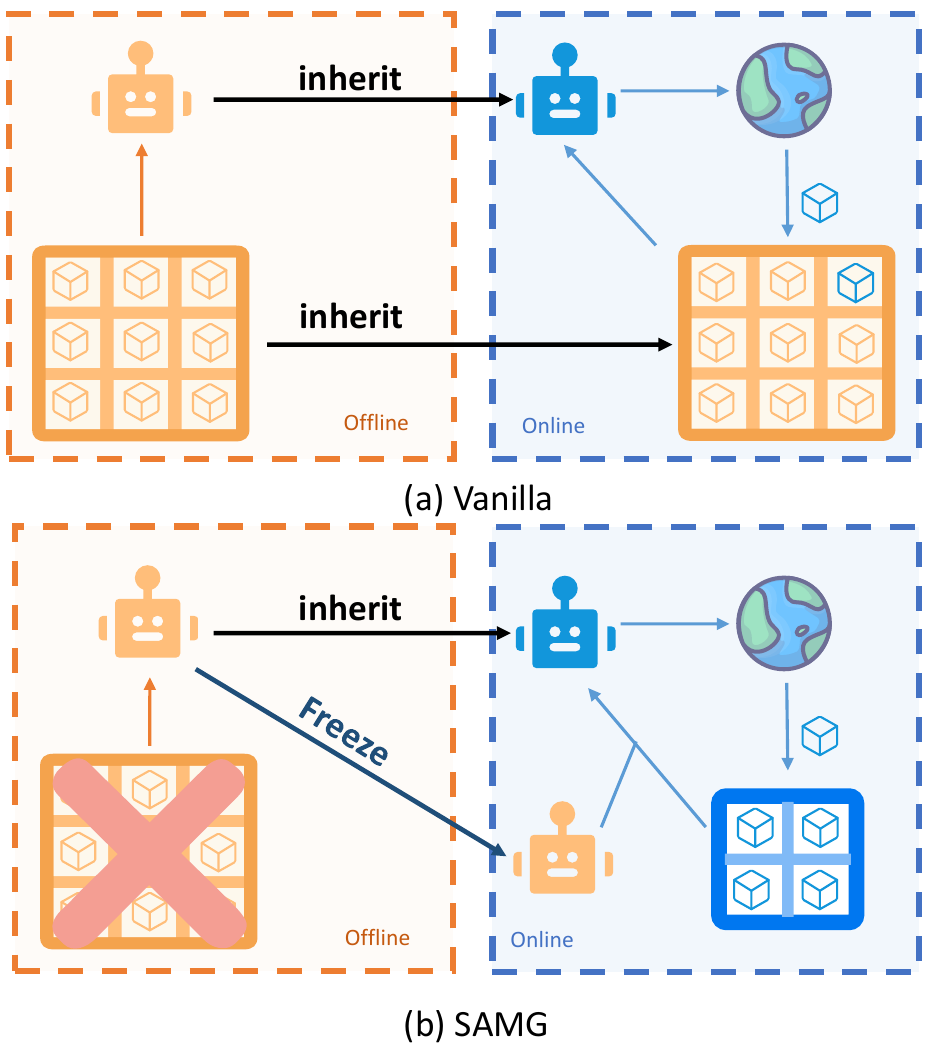} 
\caption{Compared with the vanilla O2O framework (a), the proposed \methodname\ (b) uses the frozen offline RL model with an adaptive distribution-aware model to achieve online fine-tuning with 100\% online sample rates.}
\label{fig:schematics}
\end{figure}

This compromise results in several undesirable outcomes. Training with offline data can potentially hindering algorithmic improvement given the sub-optimal nature of some offline data. Meanwhile, the inefficiency in accessing online samples limits the ability to explore and exploit novel information, making model improvement more challenging. In summary, this setting contradicts the goal of the fine-tuning process to improve algorithm performance with limited training budget.

To tackle the challenge of low online sample utilization, it is inspiring to directly leverage the offline critic, which is meticulously extracted from the offline dataset as an abstraction of the offline information. To this end, this paper introduces a novel online fine-tuning paradigm named \textbf{S}tate-\textbf{A}ction-Conditional Offline \textbf{M}odel \textbf{G}uidance (\methodname), which eliminates the reliance on offline data and achieves 100\% online sample utilization. 

\methodname\ freezes the offline pre-trained critic, which contains the offline cognition of the values given a state-action pair and offers offline guidance for online fine-tuning process. \methodname\ combines the offline critic with online target critic weighted by a state-action-conditional coefficient to provide a compound comprehension perspective. The state-action-conditional coefficient represents a class of functions that quantify the offline confidence of a given state-action pair. It is adaptively updated during training to provide accurate probability estimation. \methodname\ avoids introducing inappropriate intrinsic rewards by leveraging this probability-based mechanism. It does not affect offline algorithms and can be easily deployed on Q-function-based RL algorithms, demonstrating strong applicability.


The main contributions of this paper are summarized as follows: (1) The tedious offline data is eliminated to facilitate more effective online sample utilization. (2) The compact offline information generated by offline model is integrated to provide offline guidance. A novel class of state-action-conditional function is designed and updated to estimate the offline confidence. (3) Rigorous theoretical analysis demonstrates good convergence and lower estimation error. \methodname\ is integrated into four Q-learning-based algorithms, showcasing remarkable advantages.

\section{Preliminaries}
\label{sec:Preliminaries}
\textbf{Reinforcement Learning} task is defined as a sequential decision-making process, where an RL agent interacts with an environment modeled as a Markov Decision Process (MDP): $\mathcal{M} = (\mathcal{S}, \mathcal{A}, P, r, \gamma, \tau)$. $\mathcal{S}$ represents the state space and $\mathcal{A}$ represents the action space. $P(s' | s, a)$ denotes the unknown function of transition model and $r(s, a)$ denotes the reward model bounded by $|r(s, a)| \leq R_{max}$. $\gamma\in(0, 1)$ denotes the discount factor for future reward and $\tau$ denotes the initial state distribution. The goal of the RL agent is to acquire a policy $\pi (a | s)$ to maximize the cumulative discounted reward, defined as value function $V^\pi (s) = {\mathbb{E}_{\pi} \left[\sum_{k=0}^{\infty}\gamma ^k r(s_k, a_k) | s_0=s\right]}$ or state-action value function $Q^\pi(s,a) = {\mathbb{E}_{\pi} \left[\sum_{k=0}^{\infty}\gamma^k r(s_k, a_k) | s_0=s,a_0=a\right]}$. The training process for actor-critic algorithms alternates between policy evaluation and policy improvement phases. Policy evaluation phase maintains an estimated Q-function $Q_\theta(s, a)$ parameterized by $\theta$ and updates it by applying the Bellman operator: $\bellman^\pi Q \doteq r + \gamma P^\pi Q$, where $P^{\pi_\phi} Q(s, a) = \mathbb{E}_{s^\prime \sim P(s^\prime | s, a), a^\prime \sim \pi_\phi(a^\prime | s^\prime)} {\bigl[Q(s^\prime, a^\prime)\bigr]}$. In the policy improvement phase, the policy $\pi_\phi (a)$ is  parameterized by $\phi$ and updated to achieve higher expected returns.

\textbf{Conditional Variational Autoencoder (C-VAE)} is a generative model designed to capture complex conditional data distributions by incorporating additional information. The encoder maps the input data to the mean $z_m$ and variance $z_v$ parameters of a Gaussian distribution. Latent vector $z$ is then sampled from this estimated distribution $\mathcal{N}(z_m, z_v)$ and then fed to the decoder to reconstruct the data~\cite{kingma2014semi}. C-VAE is broadly used in offline RL to approximate the behavior policy~\cite{fujimoto19a,kumar2019stabilizing,xu2022constraints} or to estimate the state-action density~\cite{guo2023simple}.


\section{SAMG: Methodology}
\label{sec:empirical-method}
\begin{figure*}[t]
\centering
{\includegraphics[trim={220 230 230 212},clip,width=0.72\linewidth]{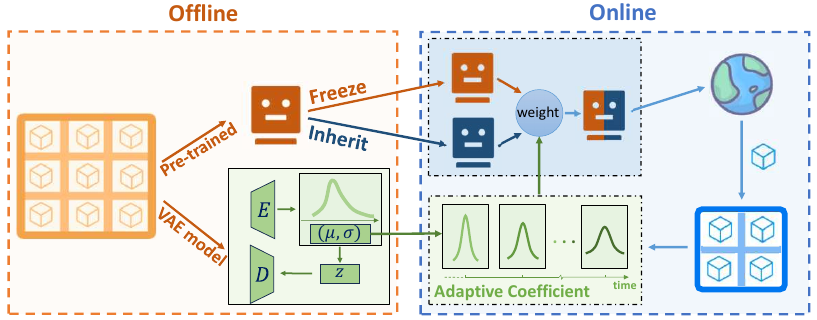}} 
\caption{\label{fig:algorithm_figure} \footnotesize{\textbf{Architecture of \methodname}. This figure illustrates the structure of \methodname, highlighting the transition from offline pre-training to online fine-tuning. It outlines key components, including offline critic, VAE model, offline-guidance technique, and adaptive coefficient.}}
\end{figure*}

In this section, we introduce the \methodname\ paradigm, which leverages the pre-trained offline model to guide the online fine-tuning process without relying on tedious offline data. This approach raises three key questions:: 
\begin{itemize}
\item[1] How can we accurately extract the concentrated information contained within the offline model?
\item[2] How can we assess the reliability of this information?
\item[3] How can we adaptively adjust the level of reliability throughout the training process?
\end{itemize}
To resolve these challenges, We propose a novel model-guidance technique and introduce an adaptive state-action-conditional coefficient, as illustrated in Fig.~\ref{fig:algorithm_figure}.
\subsection{Offline-model-guidance Paradigm}
Offline-model-guidance paradigm is designed to address Problem 1. Intuitively, the offline pre-trained value function $Q_\theta^{off} (s, a)$ of an algorithm estimates the quality of a specific state-action pair in the perspective of the offline dataset. This well-trained offline Q-network can be frozen and preserved to provide offline opinion when encountering online state-action pairs. To leverage both offline and online sights, the frozen offline Q-values are integrated with online Q-values weighted by a state-action-conditional coefficient. This approach brings several advantages: it can adaptively utilize the offline information based on its reliability and eliminate introducing undesirable intrinsic rewards, which will be discussed later. Formally, the policy evaluation equation can be obtained as follows:
\begin{equation}
    \begin{aligned}
    \label{eqn:efom_initial_equation}
    Q(s, a) =& r(s, a) + \gamma \bigl[{{\color[HTML]{0000FF}{(1-p(s, a))}} Q(s^\prime, a^\prime)} \bigr.\\
    & +\left.{\color[HTML]{0000FF}{p(s, a) Q^{off}(s^\prime, a^\prime)}} \right].
    \end{aligned}
\end{equation}
where $Q(s, a)$ represents the estimated Q-function and $p \in (0, 1)$ denotes a function class that gives a state-action-conditional coefficient and could be implemented with any reasonable form. The novel parts of the equation compared to the standard Bellman equation are marked in {\color[HTML]{0000FF}{blue}}.

\subsection{State-action-conditional Coefficient}
\label{VAE_section}
State-action-conditional coefficient is proposed to address Problem 2. Intuitively, we tend to allocate higher values to samples within the offline distribution, as these samples are well-characterized and thoroughly pre-trained. Conversely, we have limited knowledge about samples distant from the offline distribution (treated as OOD samples), so lower values are appropriate. In summary, the state-action-conditional coefficient should capture the offline confidence of online samples. Any structure that satisfies the criteria can serve as an instantiation of $p(s, a)$.


The C-VAE model is primarily adopted to instantiate the state-action-conditional coefficient in this paper. C-VAE has been widely applied in estimating the behavior policies in offline RL with its ability to model complex distributions and provide sample probability.


Nevertheless, we observe that the previous C-VAE structure suffers from posterior collapse~\cite{lucas2019understanding,wang2021posterior}, as illustrated in Appendix~\ref{apx:vae_posterior_collapse}. Posterior collapse implies that the encoder structure completely fails. The KL-divergence loss vanishes to zero for any input, and the latent output is just a standard normal distribution (0 for the mean and 1 for the variance). Thus, the decoder structure takes noise $z\sim \mathcal{N}(0, 1)$ as input and reconstructs samples all by itself. This phenomenon is highly detrimental in our setting because we need the latent output to calculate the state-action-conditional coefficient but the structure now fails to operate.

To mitigate the adverse impacts of posterior collapse, we extend the variational conditional information to include state-action pairs and reconstruct the next state from the decoder. This approach complicates the modeling process and develops a state-action-conditional structure. Additionally, we employ the KL-annealing technique~\cite{bowman2015generating}, as detailed in Appendix~\ref{apx:vae_implementations}. Formally, \methodname\ adopts the following evidence lower bound (ELBO) on the offline dataset to get the state-action-conditional C-VAE:
\begin{equation}
    \begin{aligned}
    \label{eqn:vae_elbo}
    \max \limits_{\psi _1, \psi _2} \mathbb{E}_{z \sim \text{Enc}_{\psi_1}}& \bigl[\log \text{Dec}_{\psi_2} (s^\prime | z, s, a)\bigr]\\
&- \beta D_{KL} \bigr[{\text{Enc}_{\psi_1}(z|s, a)} || {\text{Dec}_{\psi_2}(z)}\bigr]
    \end{aligned}    
\end{equation}
where $\text{Enc}_{\psi_1}(z|s, a)$ and $ \text{Dec}_{\psi_2} (s^\prime | z, s, a)$ represent the encoder and decoder structure respectively; $\text{Dec}_{\psi_2}(z)$ denotes the prior distribution of the encoder; and $D_{KL}\left[p||q\right]$ denotes the KL divergence. The former error term denotes the reconstruction loss while the latter denotes the KL divergence between the encoder distribution and the prior distribution.


\subsection{Coefficient Generation and Adaptive Updates} 
\subsubsection*{Posterior Distribution Generation}
To evaluate the effectiveness of improved C-VAE structure, we input the offline dataset to the trained C-VAE model and record the mean and variance values of encoder output, as illustrated in the Appendix~\ref{apx:practical_VAE_distribution}. The results indicate that posterior collapse is significantly alleviated. However, it is still unreliable to directly utilize the latent information $z$ because it is sampled from the narrow normal distribution and the sampling randomness overshadows the distribution information. To address this issue, we use the deterministic outputs $(z_m, z_v)$ in place of less reliable $z$. Since the statistical distribution of the encoder outputs $(z_m, z_v)$ still resembles a normal distribution, we fit these outputs to the corresponding normal distribution to represent the offline dataset. Specifically, the mean $z_m$ is modeled as $\mathcal{N}_M (\mu_m, \sigma_m)$ and the standard $z_v$ is modeled as $\mathcal{N}_S (\mu_v, \sigma_v)$.

Then, the offline coefficient of a given sample $(x_m, x_v)$ can be devised as the probability falling within a distance less than $|z_m-\mu_m|$ for mean and $|z_v-\mu_v|$ for variance. The cumulative distribution function $F_X(x)$ is leveraged to get the probability of an online sample aligning with the offline distribution. The intermediate probability can be obtained:
\begin{equation}
    \begin{split}
    \label{eqn:init_vae_probability}
    \begin{aligned}
        p^{int}(s, a)=\omega &|F_{Z_m}(z_m)-F_{Z_m}(2\mu_m- z_m)|+\\
        &(1 - \omega) |F_{Z_v}(z_v)-F_{Z_v}(2\mu_v- z_v)|
    \end{aligned}
    \end{split}
\end{equation} 
where $\omega$ is the weight of mean and standard. This weight is set to 1 because the estimation error of the mean is significantly smaller than that of the standard.

Moreover, in cases where the sample diverges notably from the offline distribution, the information about the sample is unknown so the probability is manually set to be 0, with a hyperparameter $p^{vae}_m$ introduced. The eventual  equation to calculate the probability $p^{off}$ of a given sample $(z_m, z_v)$ is illustrated below:
\begin{equation}
    \begin{split}
    \label{eqn:vae_probability}
    p^{vae}(s, a)=\left\{
    \begin{aligned}
        &p^{int}(s, a), &p^{int}(s, a)\geq p^{vae}_m\\
        &0, &p^{int}(s, a) < p^{vae}_m   
    \end{aligned}
    \right.
    \end{split}
\end{equation} 




By integrating the C-VAE form state-action-conditional coefficients into Eq. (\ref{eqn:efom_initial_equation}), the following practical updating equation can be obtained:
\begin{equation}
\begin{aligned}
\label{eqn:efom_final_equation}
Q(s, a) =& r(s, a) + \gamma \left[{\color[HTML]{0000FF}(1-p^{off}(s, a))} Q(s^\prime, a^\prime)\right.\\
&+ \left.{\color[HTML]{0000FF}{p^{off}(s, a) Q^{off}(s^\prime, a^\prime)}}\right].
\end{aligned}    
\end{equation}

\textbf{Comprehension of state-action-conditional coefficient.} The state-action-conditional coefficient module actually attempts to depict the characteristics of the complex distribution represented by the offline dataset. Considering the high-dimensional and continuous property of the state-action data, it is challenging to directly extract the probability characteristics from the state-action pair. Therefore, it is required to perform dimension reduction of the distribution of the offline dataset, i.e., the complex distribution is simplified to be a normal distribution.  

\subsubsection*{Adaptive Coefficient Updates}
During the online training process, the fine-tuned model gradually adapts to the environment and learns to handle some of the OOD samples. However, the offline guidance and state-action-conditional coefficient remain fixed, which becomes more and more unsuitable for probability generation. To address this issue (Problem 3), adaptive coefficient updates are proposed. At fixed intervals, OOD samples from the current period are collected, and these OOD samples with the lowest offline Q-loss are identified and reclassified as mastered OOD samples. They are then used to retrain the VAE model, allowing the VAE coefficient to adapt to the online fine-tuning process. Meanwhile, the offline pre-trained model is updated to the Q-function at these specific intervals. Refer to Appendix~\ref{apx:adaptive_vae_coefficient} for complete implementation.

\section{Analysis of \methodname}

\subsection{Intuitive analysis of \methodname}
\label{sec:empirical}

\textbf{Intrinsic Reward Analysis} highlights the importance of probability-based coefficient. Specifically, Eq. (\ref{eqn:efom_final_equation}) can be derived as below:
\begin{equation}
\begin{aligned}
\label{eqn:intrinsic_reward_equation}
Q(s, a) = \bigl[r(s, a) + r^{in}(s, a)\bigr]+ \gamma Q(s^\prime, a^\prime).
\end{aligned} 
\end{equation}
where $r^{in}(s, a)=\gamma p(s, a) (Q^{off}(s^\prime, a^\prime) - Q(s^\prime, a^\prime))$. Eq. (\ref{eqn:intrinsic_reward_equation}) indicates that the induced offline information could be treated as intrinsic reward.

Previous work has revealed that intrinsic reward may cause training instability or even algorithm degradation~\cite{chen2022redeeming,mcinroe2024planning}. However, the intrinsic reward form of \methodname\ is reasonable and stable thanks to the probability-shape coefficient. 

Specifically, the intrinsic reward term describes the difference between offline and online Q-values, weighted by the state-action-conditional coefficient. It can be analyzed in two scenarios. Firstly, if the state-action pair lies within the offline distribution, where the offline Q-value is well trained and the state-action-conditional coefficient $\alpha$ is significant. In this case, this term suggests that higher offline Q-values correspond to higher potential returns. Hence, it encourages exploring state-action pairs with higher performance. Conversely, if the state-action pair falls outside the offline distribution, where the offline Q-value may be erroneously estimated. This term becomes negligible or is even set to zero, as specified in Eq. (\ref{eqn:vae_probability}). Therefore, it can filter out inaccurate and questionable information. In summary, \methodname\ is able to properly retain the offline knowledge without introducing inappropriate intrinsic rewards.



\subsection{Theoretical analysis of \methodname}
\label{sec:theory}
In this section, we adopt the temporal difference paradigm~\cite{sutton1988learning,haarnoja2018soft} in the tabular setting and prove that Eq. (\ref{eqn:efom_final_equation}) still converges to the same optimality, even with an extra term induced. For the theoretical tools, \methodname\ gets rid of the offline dataset and therefore diverges from the hybrid realm of Song et al.~\cite{song2023hybrid} and offline RL scope limited by the dataset, but aligns with traditional online RL algorithms~\cite{convergence1994,thomas2014bias,haarnoja2018soft}. 

\textbf{Contraction Property} is considered and proven to still hold since the standard Bellman operator is broken~\cite{keeler1969theorem}. The related theorem and detailed proof can be found in Appendix~\ref{Apx:contraction_mapping_property}. 

\textbf{Convergence Optimality.} Formally, the iterative TD updating form of Eq. (\ref{eqn:efom_final_equation}) is demonstrated as below:
\begin{equation}
    \label{eqn:theoretical_iterative_expression}
    \begin{split}
        &Q_{k+1} (s, a)-Q_{k} (s, a)=\\
        &\alpha_k (s, a) \left[Q_k(s, a) -\left( r_{k+1} + \gamma Q_k (s_{k + 1}, a_{k + 1})\right) \right]-\\
        &\alpha_k (s, a)\gamma p(s, a)\left( Q^{off}(s_{k + 1}, a_{k + 1}) - Q_k(s_{k + 1}, a_{k + 1})\right)
    \end{split}
\end{equation}
where the estimated state-value function at time-step $k$ for given $(s, a)$ pair is denoted as $Q_k(s, a)$. The learning rate at time-step $k$ is represented as $\alpha_k$. For simplicity, the state-action conditional coefficient $p^{off}(s, a)$ and threshold $p^{off}_m(s,a)$ are denoted as $p(s, a)$ and $p_m(s, a)$, respectively. 
 




The proof pipeline involves deriving the incremental form of \methodname\, then proving that the incremental form satisfies Dvoretzky's Theorem~\cite{dvoretzky1959theorem} and finally arriving at the following theorem. Details refer to Appendix~\ref{Apx:ConvergenceAnalysis}.

\begin{theorem}[Convergence property of \methodname]
\label{thm:convergence_optimality}
For a given policy $\pi$, by the TD updating paradigm, $Q_k(s, a)$ of \methodname\ converges almost surely to $Q^{\pi}(s, a)$ as $k\to \infty$ for all $s \in \mathcal{S}$ and $a \in \mathcal{A}$ if $\sum_k \alpha_k(s, a)=\infty$ and $\sum_k \alpha_k^2(s, a) < \infty$ for all $s \in \mathcal{S}$ and $a \in \mathcal{A}$. 
\end{theorem}

\textbf{Convergence Speed.} Moreover, the specific expression for the contraction coefficient is proven as follows, illustrating the faster convergence speed of \methodname. See Appendix~\ref{Apx:convergence_speed} for further details.
\begin{theorem}[Convergence speed of \methodname]
\label{thm:convergence_speed}
The Bellman operator of \methodname\ satisfies the contraction property $\big\Vert \bellman(x) - \bellman(y)\big\Vert \leq \gamma_o \big\Vert x, y\big\Vert$ for all $x, y \in \mathcal{Q}$. $\mathcal{Q}$ represents the Q function space. The contraction coefficient of \methodname\ $\gamma_o (s, a)$ is bounded above by the following expression:
\begin{align}
    \left\{
    \begin{aligned}
    \label{eqn:TD_error_term}
        &\left(1-p(s, a)\right)\gamma + \gamma \gamma_{\mathcal{F}}p(s, a)C,&p(s, a)\geq p_m\\
    &\gamma, &p(s, a)<p_m
    \end{aligned}
    \right.
\end{align}
where $C=\big\Vert \Delta_{off}(s, a) \big\Vert_\infty \big/ \big\Vert\Delta_k(s, a) \big\Vert_\infty$ denotes the ratio of the offline and online suboptimality bounds, $\big\Vert \Delta_{off}(s, a) \big\Vert_\infty$ denotes the offline suboptimality bound $\left[V^*(s) - V^{\pi_{off}}(s)\right]$, $\big\Vert\Delta_k(s, a) \big\Vert_\infty$ denotes the suboptimality bound of the k-th iteration of online fine-tuning and $0<\gamma_{\mathcal{F}}<1$ denotes the convergence coefficient of offline algorithm class $\mathcal{F}$.
\end{theorem}
The upper equation of Theorem~\ref{thm:convergence_speed} holds for in-distribution samples, which are well mastered by the offline model. Therefore, the offline suboptimality bound is substantially tighter compared to the online bound. This illustrates that the offline model guidance significantly accelerates the online fine-tuning process by providing more accurate estimations for in-distribution samples. Specifically, for these samples, the convergence speed depends on the offline confidence implied by $p(s, a)$, i.e., a higher $p(s, a)$ indicates a higher degree of offline-ness, corresponding to a smaller error term constrained by the term $(1-p(s, a))$ and ensuring faster convergence. For the OOD samples, the algorithm degenerates into the traditional algorithm because $p(s, a)$ is set to zero as defined in Eq. (\ref{eqn:vae_probability}). This theoretical result is highly consistent with the analysis of the expected performance as stated in Section~\ref{sec:empirical}. Furthermore, it indicates that the extent of algorithm improvement is influenced by the sample coverage rate of the offline dataset. Specifically, the offline guidance is more reliable with more complex sample coverage, whereas the guidance is constrained with limited sample diversity. 
\section{Experimental Results}
\label{sec:experiments}

Our experimental evaluations focus on the performance of \methodname\ during the online fine-tuning process based on three state-of-the-art algorithms within D4RL~\cite{fu2020d4rl}, covering diverse environments and task complexities, as detailed below. 



{
\setlength\fboxsep{2pt}
\begin{table*}[ht!]
\centering
\begin{threeparttable}
  \small
  \caption{\textbf{Performance comparison.}
The D4RL normalized score~\cite{fu2020d4rl} is evaluated of standard base algorithms (including CQL~\cite{kumar2020conservative}, IQL~\cite{kostrikov2022offline} and AWAC~\cite{nair2020accelerating}, denoted as ``Vanilla") in comparison to the base algorithms augmented with \methodname\ (referred to as ``Ours"), as well as three baselines (TD3BC~\cite{chen2020bail}, SPOT~\cite{wu2022supported}, Cal\_QL~\cite{nakamoto2023calql} and EDIS~\cite{liu2024energyguided}).
The superior scores are highlighted in \colorbox{myblue}{blue}. The result is the average normalized score of 5 random seeds $\pm$ (standard deviation). 
%
  }
  \label{tab:vinilla_vs_OMDA}
  \centering
  \renewcommand{\tabcolsep}{3.2pt} 
  \begin{tabular}{l|cccccc|cccc} 
    \specialrule{0.12em}{0pt}{0pt}
	\multirow{2}{*}{Dataset\tnote{1}}
	& \multicolumn{2}{c}{CQL} 
	& \multicolumn{2}{c}{AWAC} 
	& \multicolumn{2}{c|}{IQL}
        & \multirow{2}{*}{TD3BC}
        & \multirow{2}{*}{SPOT}
        & \multirow{2}{*}{Cal\_QL}
        & \multirow{2}{*}{EDIS}
	\\ 
	\cline{2-7}
	& Vanilla & Ours
	& Vanilla & Ours
	& Vanilla & Ours
	\\ \hline
    Hopper-mr
    & 100.6\tiny{(1.8)} & \colorbox{myblue}{103.7}\tiny{(1.3)} & 99.4\tiny{(1.3)} & \colorbox{myblue}{108.3}\tiny{(0.2)} & 86.2\tiny{(16.1)} & \colorbox{myblue}{100.4}\tiny{(0.9)}
    &64.4\tiny{(21.5)}&68.0\tiny{(11.2)} & {80.9}\tiny{(38.2)} &83.0\tiny{(26.8)}
    \\
    Hopper-m
     & 60.2\tiny{(2.7)} & \colorbox{myblue}{88.3}\tiny{(6.0)} & 88.2\tiny{(14.6)} & \colorbox{myblue}{102.5}\tiny{(1.8)} & {62.1}\tiny{(7.4)} & \colorbox{myblue}{68.4}\tiny{(2.9)} &
     66.4\tiny{(3.5)}&54.6\tiny{(7.1)}& 78.1\tiny{(8.7)} &30.1\tiny{(8.9)}
    \\
    Hopper-me
     & 110.8\tiny{(1.0)} & \colorbox{myblue}{113.0}\tiny{(0.3)} & 101.9\tiny{(20.5)}  & \colorbox{myblue}{112.8}\tiny{(7.2)} & 103.5\tiny{(8.7)} & \colorbox{myblue}{108.1}\tiny{(3.1)}&
     101.2\tiny{(9.1)}&82.6\tiny{(11.5)} & 109.1\tiny{(0.2)} & 78.4\tiny{(3.5)}
    \\ 
     Half-mr
    & 48.0\tiny{(0.5)} & \colorbox{myblue}{57.8}\tiny{(1.7)} & 48.9\tiny{(1.1)} & \colorbox{myblue}{62.8}\tiny{(3.3)} & 45.1\tiny{(0.6)} & \colorbox{myblue}{49.6}\tiny{(1.0)} & {44.8}\tiny{(0.6)}
    &42.4 \tiny{(3.7)}&51.6\tiny{(0.8)}&82.9\tiny{(1.2)}
    \\
    Half-m
     & 47.6\tiny{(0.2)} & \colorbox{myblue}{59.0}\tiny{(0.7)} & 54.2\tiny{(1.1)} & \colorbox{myblue}{69.5}\tiny{(0.9)} & {49.3}\tiny{(0.1)} & \colorbox{myblue}{62.5}\tiny{(1.5)} & {48.1}\tiny{(0.2)}&
     45.9\tiny{(2.4)}&63.2\tiny{(2.5)}& 66.4\tiny{(11.7)}
    \\ 
    Half-me
     & 95.2\tiny{(1.0)} & \colorbox{myblue}{97.2}\tiny{(0.8)} & 94.8\tiny{(1.3)} & \colorbox{myblue}{96.7}\tiny{(1.0)} & \colorbox{myblue}{91.6}\tiny{(0.9)} & {82.3}\tiny{(11.7)} & {90.8}\tiny{(6.0)}&
     87.4\tiny{(7.4)}&95.6\tiny{(4.3)}&90.2\tiny{(1.4)}
     \\
      Walker2d-mr
    & 82.7\tiny{(0.7)} & \colorbox{myblue}{88.4}\tiny{(5.0)} & 93.8\tiny{(3.4)} & \colorbox{myblue}{120.1}\tiny{(3.1)} & 87.1\tiny{(3.3)} & \colorbox{myblue}{99.5}\tiny{(2.4)} & {85.6}\tiny{(4.0)}
    &69.2 \tiny{(6.2)}&97.1\tiny{(2.5)}&46.9\tiny{(23.6)}
    \\
    Walker2d-m
     & 60.2\tiny{(2.7} & \colorbox{myblue}{82.9}\tiny{(1.8)} & 87.8\tiny{(0.8)} & \colorbox{myblue}{103.6}\tiny{(4.5)} & {83.4}\tiny{(1.6)} & \colorbox{myblue}{88.6}\tiny{(4.7)} & {82.7}\tiny{(4.8)}&
     79.5\tiny{(2.4)}&83.6\tiny{(0.8)}&76.2\tiny{(16.7)}
    \\
    Walker2d-me
     & 109.5\tiny{(0.5)} & \colorbox{myblue}{112.5}\tiny{(0.7)} & 112.7\tiny{(0.9)} & \colorbox{myblue}{129.2}\tiny{(3.6)} & 113.6\tiny{(1.1)} & \colorbox{myblue}{116.3}\tiny{(3.7)} & {110.0}\tiny{(0.4)}&
     87.8\tiny{(3.9)}&110.7\tiny{(0.4)}&107.9\tiny{(10.3)}
     \\
    \hline
    Antmaze-u
     & 92.0\tiny{(1.7)} & \colorbox{myblue}{97.0}\tiny{(1.4)} & 70.0\tiny{(40.4)} & \colorbox{myblue}{87.0}\tiny{(13.2)} & 83.3\tiny{(6.1)} & \colorbox{myblue}{94.0}\tiny{(1.2)}&
    70.8\tiny{(39.2)}&30.8\tiny{(12.9)} & 96.8\tiny{(0.4)} &95.0\tiny{(7.0)}
    \\ 
    Antmaze-ud 
    & {58.0}\tiny{(32.0)} & \colorbox{myblue}{62.3}\tiny{(12.4)} & 15.0\tiny{(35.3)} & \colorbox{myblue}{75.0}\tiny{(7.0)} & 33.3\tiny{(4.4)} & \colorbox{myblue}{77.8}\tiny{(0.8)} &
    44.8\tiny{(11.6)}&44.8\tiny{(6.5)}& 63.8\tiny{(43.4)} & 72.5\tiny{(32.5)}
    \\ 
    Antmaze-md
    & 82.4\tiny{(2.2)} & \colorbox{myblue}{89.1}\tiny{(3.5)} & 0.0\tiny{(0.0)} & 0.0\tiny{(0.0)} & 76.5\tiny{(5.4)} & \colorbox{myblue}{96.5}\tiny{(1.9)} &  
    0.3\tiny{(0.4)}&36.3\tiny{(11.0)} & {93.4}\tiny{(3.6)} &82.4\tiny{(4.8)}
    \\ 
    Antmaze-mp
    & 85.6\tiny{(6.6)} & \colorbox{myblue}{86.1}\tiny{(1.1)} & 0.0\tiny{(0.0)} & 0.0\tiny{(0.0)} & 76.2\tiny{(4.6)} & \colorbox{myblue}{95.2}\tiny{(1.6)} & 
    0.3\tiny{(0.4)}&38.5\tiny{(8.7)}&{94.0}\tiny{(2.2)}&60.0\tiny{(51.9)}
    \\ 
    Antmaze-ld
    & 62.8\tiny{(7.4)} & \colorbox{myblue}{63.9}\tiny{(5.9)} & 0.0\tiny{(0.0)} & 0.0\tiny{(0.0)} & 45.4\tiny{(7.7)} & \colorbox{myblue}{81.5}\tiny{(7.9)} &0.0\tiny{(0.0)}&0.0\tiny{(0.0)}&78.8\tiny{(5.8)}&32.5\tiny{(15.0)}
    \\
    Antmaze-lp
     & 55.0\tiny{(8.4)} & \colorbox{myblue}{60.7}\tiny{(1.5)} & 0.0\tiny{(0.0)} & 0.0\tiny{(0.0)} & 48.8\tiny{(7.7)} & \colorbox{myblue}{74.8}\tiny{(8.4)} &0.0\tiny{(0.0)}&0.0\tiny{(0.0)}&{73.0}\tiny{(19.4)}&35.0\tiny{(17.3)}
    \\ 
    \hline
    Pen-b
    & 90.0\tiny{(4.6)} & \colorbox{myblue}{96.2}\tiny{(4.0)} & 63.3\tiny{(39.7)} & \colorbox{myblue}{70.1}\tiny{(26.6)} & 86.7\tiny{(24.6)} & \colorbox{myblue}{106.0}\tiny{(27.8)} &6.4\tiny{(4.37)}&2.80\tiny{(11.82)}&{-0.03}\tiny{(4.10)}&8.99\tiny{(17.18)}
    \\ 
    Door-b
    & {-0.34}\tiny{(0.01)} & \colorbox{myblue}{70.8}\tiny{(2.8)} & 0.00\tiny{(0.01)} & \colorbox{myblue}{7.29}\tiny{(3.14)} & -0.06\tiny{(0.03)} & \colorbox{myblue}{13.08}\tiny{(3.20)} & 
    -0.32\tiny{(0.01)}&-0.16\tiny{(0.05)}&-0.33\tiny{(0.01)}&0.09\tiny{(0.03)}
    \\ 
    Relocate-b
     & -0.28\tiny{(0.12)} & \colorbox{myblue}{75.0}\tiny{(2.1)} & -8.84\tiny{(1.22)} & \colorbox{myblue}{-7.82}\tiny{(0.93)} & -0.01\tiny{(0.04)} & \colorbox{myblue}{0.24}\tiny{(0.03)} &-0.21\tiny{(0.01)}&-0.14\tiny{(0.10)}&{-0.31}\tiny{(0.03)}&-0.34\tiny{(0.02)}
    \\  
    \specialrule{0.12em}{0pt}{0pt}
  \end{tabular}
  \begin{tablenotes}
            \item[1] Half: HalfCheetah, mr: medium-replay, me: medium-expert, d: diverse, p: play, u: umaze, ud: umaze-diverse, md: medium-diverse, mp: medium-play, ld: large-diverse, lp: large-play, b: binary.
    \end{tablenotes}
\end{threeparttable}
\end{table*}
}

\begin{figure*}[t]
\centering
\begin{minipage}[t]{0.32\textwidth}
\centering
    \includegraphics[width=1.0\columnwidth]{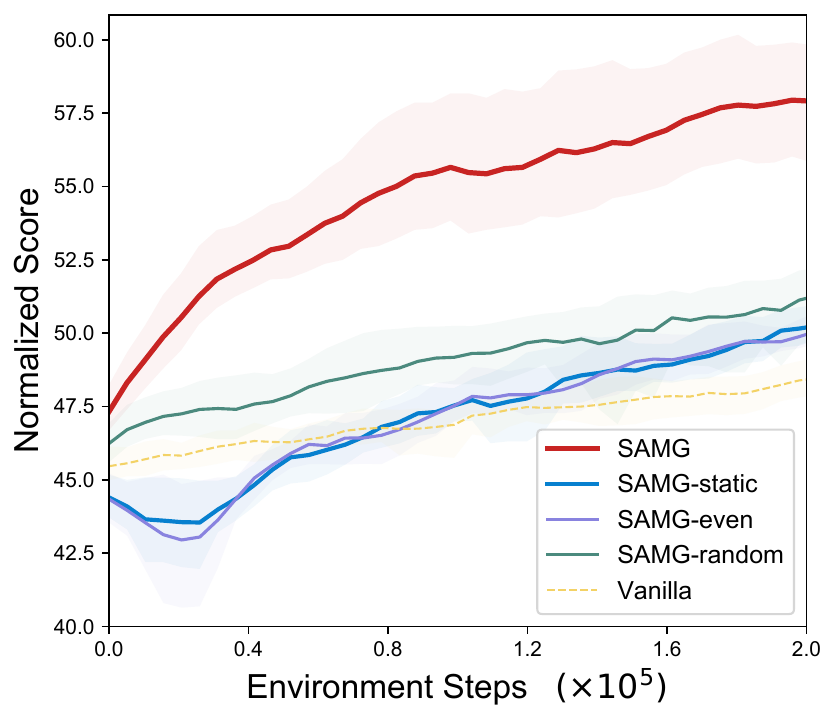}
    \caption{Ablation analysis of the state-action-conditional coefficient.}
    \label{fig:abalation study of distribution-aware weight}
\end{minipage}
\ 
\begin{minipage}[t]{0.32\textwidth}
\centering
    \includegraphics[width=1.0\columnwidth]{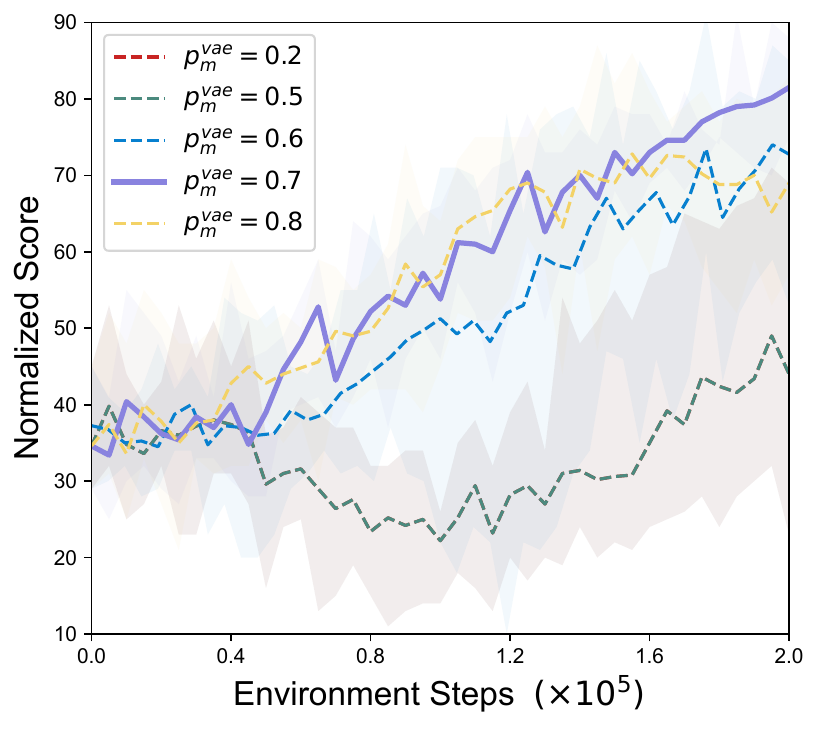}
    \caption{Sensitivity test for the coefficient threshold $p^{vae}_{m}$.}
    \label{fig:sensitivity of the p_om}
\end{minipage}
\ 
\begin{minipage}[t]{0.32\textwidth}
\centering
\includegraphics[width=1.0\columnwidth]{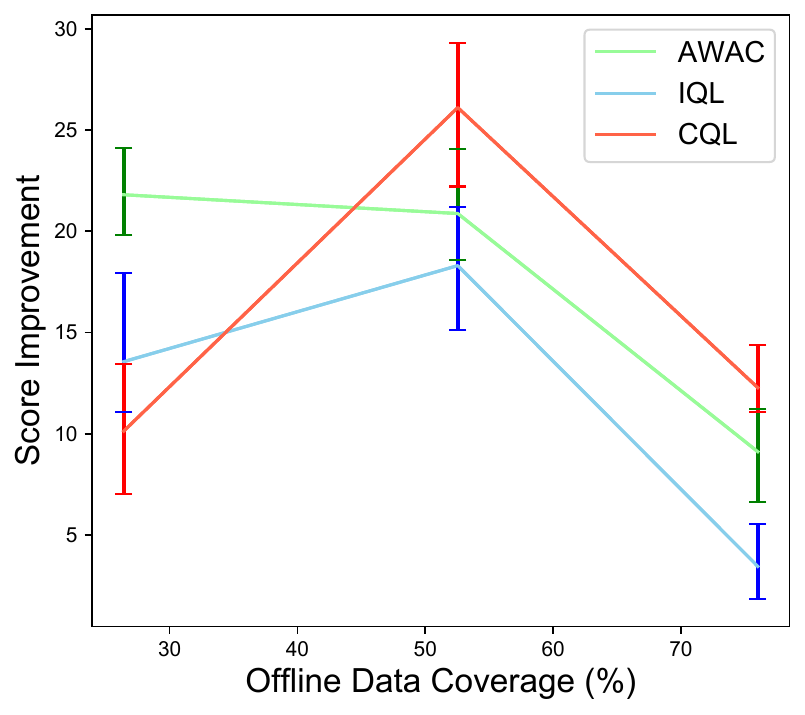}
    \caption{Average improvement of the normalized score over the offline data coverage.}
    \label{fig:offlinecoverage}
\end{minipage}
\end{figure*}

\textbf{Baselines.} (i) \textbf{\methodname\ algorithms}, the \methodname\ paradigm is constructed on a variety of state-of-the-art O2O RL algorithms, including CQL~\cite{kumar2020conservative} and AWAC~\cite{nair2020accelerating}, IQL~\cite{kostrikov2022offline}. (ii) \textbf{O2O RL algorithms}, we implement the aforementioned O2O RL algorithms (CQL, AWAC and IQL). We also implement SPOT~\cite{wu2022supported} (iii) \textbf{Hybrid RL}, we implement sota hybrid-RL-based algorithms, including Cal\_QL~\cite{nakamoto2023calql} and EDIS~\cite{liu2024energyguided}. (iv) \textbf{Behavior Cloning (BC)}, we implement Behavior cloning based algorithm TD3+BC~\cite{fujimoto2021minimalist}. All algorithms are implemented based on CORL library~\cite{tarasov2024corl} with implementation details in Appendix~\ref{Apx:implementations_on_baselines}. To ensure a fair comparison, all algorithms are pre-trained offline for \textbf{1M iterations} followed by \textbf{200k iterations} of online fine-tuning.

\textbf{Benchmark tasks.}  We evaluate \methodname\ and the baselines across multiple benchmark tasks: \textbf{(1)} The Mujuco locomotion tasks~\cite{fu2020d4rl}, including three different kinds of environments ({HalfCheetah, Hopper, Walker2d}) where robots are manipulated to complete various tasks on three different levels of datasets. \textbf{(2)} The {AntMaze} tasks that an ``Ant'' robot is controlled to explore and navigate to random goal locations in six levels of environments. \textbf{(3)} The Adroit tasks include Pen, Door, and Relocate environments. Details are stated in Appendix~\ref{apx:dataset_description}.

\subsection{Empirical Results}
\label{sec:empirical_results}
The normalized scores of the vanilla algorithms with and without \methodname\ integrated are shown in Table~\ref{tab:vinilla_vs_OMDA}.
\methodname\ consistently outperforms the vanilla algorithms in the majority of environments, illustrating the superiority of \methodname. \methodname\ converges significantly faster than vanilla algorithms and can achieve higher performance. 
Notably, \methodname\ achieves the best performance with the simpler algorithm AWAC, while delivering substantial improvements with other algorithms. The reason for this counter-intuitive phenomenon is discussed in Appendix~\ref{Apx:conter_intuition_reason}, which just illustrates the effectiveness of \methodname. We present the cumulative regrets on Antmaze in Appendix~\ref{APx:cumulative_regret}, further demonstrating the outstanding online sample efficiency of \methodname.

Although \methodname\ performs well in most environments, it is still worthwhile to notice \methodname\ may occasionally behave unsatisfactory (e.g., IQL-\methodname\ on {HalfCheetah-medium-expert} task). We discuss in Appendix~\ref{Apx:unsatisfactory_performance} that this exception is caused by the unique property of the environment rather than the defect of \methodname. We notice that the AWAC algorithm performs poorly in the Antmaze environment, resulting in \methodname\ struggling to initiate. This is because AWAC is relatively simple and not competent for the complex task of Antmaze; it is an inherent limitation of AWAC, rather than an issue with  \methodname. 


\subsection{Ablation Analysis of the Coefficient}
The state-action-conditional coefficient $p(s, a)$, instantiated as $p^{vae}(s, a)$, estimates the offline degree for a given $(s, a)$ pair and is adaptively updated during training. To demonstrate the impact of this adaptive state-action-conditional coefficient, we compare several different architectures on the environment HalfCheetah with CQL-\methodname, including: (i) adopted \methodname\ setting (denoted as \methodname), (ii) static state-action-conditional coefficient (which means the VAE model is fixed once pre-trained offline, denoted as \methodname-static) (iii) 1/2 for online and 1/2 for offline (denoted as \methodname-even), (iv) randomly generated probability (denoted as \methodname-random), (v) the vanilla RL algorithms (denoted as Vanilla). The results are illustrated in Fig.~\ref{fig:abalation study of distribution-aware weight}.

\methodname\ shows consistent and significant improvement compared to other settings. Casual selections of C-VAE (\methodname-even and \methodname-random) exhibit notably inferior algorithm performance during the initial training phase, demonstrating the effectiveness of state-action-conditional coefficient structure. However, they catch up with and surpass the performance of the vanilla algorithms, highlighting the advantage of \methodname\ paradigm and offline information. \methodname\ improves over the \methodname-static and \methodname-even algorithms by \textbf{15.3\% and 21.8\%} respectively. 


Furthermore, we provide another possible implementation technique $p^{emb}$ given that $p(s, a)$ represents a function class. Refer to Appendix~\ref{apx:embedding_network} for complete details.

\subsection{Sensitivity Analysis of Probability Threshold}
\label{sec:p_om_sensitivity}
It is a crucial hyperparameter that $p^{vae}_{m}$ holds the lower threshold of OOD samples. We evaluate the sensitivity of $p^{vae}_{m}$ on environment Antmaze with IQL-\methodname\ across a range of numbers, including 0.2, 0.5, 0.6, 0.7 (chosen value), and 0.8. The results are shown in Fig.~\ref{fig:sensitivity of the p_om}. A slight change in the value of $p^{vae}_{m}$ (corresponding to 0.6 and 0.8) leads to a minor decrease in algorithm performance, but the impact is not significant. If the value is too large, the algorithm reverts to the vanilla version; if it is too small, it introduces incorrect offline critic perceptions of OOD samples, resulting in a performance decline. Furthermore, when $p^{vae}_{m}$ is small enough, all samples are regarded as OOD samples and updated solely based on offline knowledge, causing a significant drop in algorithm performance, which is also identical across different $p^{vae}_{m}$, as evidenced by the overlapping curves for 0.2 and 0.5 in Fig.~\ref{fig:sensitivity of the p_om}.



\subsection{Further Comparisons with Hybrid RL Algorithms}
\label{sec:comparisons_with_CAL_QL}
We compare \methodname\ with Hybrid RL based algorithms in Table~\ref{tab:vinilla_vs_OMDA}. A natural problem arises: would the algorithm performance improve if the hybrid RL setting were replaced with offline-guidance setting? To illustrate this problem, we modify the Cal\_QL algorithm with \methodname\ setting. Implementation details can be found in Appendix~\ref{apx:compare_hybrid_rl}. The results, presented in Table~\ref{tab:algorithm_comparison_CAL_QL}, indicate that \methodname\ still outperforms Cal\_QL, demonstrating its superiority. As for EDIS algorithm, it relies heavily on the offline dataset, making it impractical to adapt to offline-guidance setting.

{
\setlength\fboxsep{2pt}
\begin{table*}[ht]
\centering
\begin{threeparttable}
  \small
  \caption{\textbf{Algorithm performance of Cal\_QL and \methodname.}
The algorithms performance of Cal\_QL compared to \methodname\ integrated Cal\_QL algorithms. The result is the average normalized score of 5 random seeds $\pm$ (standard deviation).
  }
  \label{tab:algorithm_comparison_CAL_QL}
  \centering
  \renewcommand{\tabcolsep}{4.2pt} 
  \begin{tabular}{l|ccccccccc} 
    \specialrule{0.12em}{0pt}{0pt}
  Algorithms&Hopper-mr&Hopper-m&Hopper-me&Half-mr&Half-m&Half-me&Walk-mr&Walk-m&Walk-me
    \\ \hline
    Cal\_QL
     & 80.9\tiny{(38.2)} & {78.1}\tiny{(8.7)} & 109.1\tiny{(0.2)} & {51.6}\tiny{(0.8)} & 63.2\tiny{(2.5)} & {95.6}\tiny{(4.3)}& 97.1\tiny{(2.5)} & 83.6\tiny{(0.8)}&  110.7\tiny{(0.4)}
    \\ 
    \methodname
    & {101.6}\tiny{(0.6)} & 99.8\tiny{(2.1)} & 111.7\tiny{(0.6)} & {56.4}\tiny{(1.5)} & 65.1\tiny{(1.0)} & {96.3}\tiny{(0.5)} & 101.2\tiny{(0.1)} & 97.8\tiny{(1.7)} & 112.3\tiny{(0.8)} 
    \\
    \specialrule{0.12em}{0pt}{0pt}
    &Ant-u&Ant-ud&Ant-mp&Ant-md&Ant-lp&Ant-ld&Pen-b&Door-b&Relocate-b
    \\ \hline
    Cal\_QL
     & 96.8\tiny{(0.4)} & {63.8}\tiny{(43.4)} & 93.4\tiny{(3.6)} & 94.0\tiny{(2.2)} & 78.8\tiny{(5.8)} & {73.0}\tiny{(19.4)}&-0.03\tiny{(4.10)}&-0.33\tiny{(0.01)}& -0.31\tiny{(0.03)}
    \\ 
    \methodname
    & {99.0}\tiny{(1.0)} & 66.4\tiny{(23.0)} & 91.6\tiny{(2.4)} & {96.0}\tiny{(1.6)} & 79.8\tiny{(2.4)} & {72.4}\tiny{(11.2)} &5.01\tiny{(12.01)}&1.35\tiny{(0.57)}& 3.89\tiny{(0.34)}
    \\ 
    \specialrule{0.12em}{0pt}{0pt}
  \end{tabular}
\end{threeparttable}
\end{table*}
}

\subsection{Does \methodname\ Rely on Offline Data Coverage?}
This part aims to showcase the relationship between algorithm performance improvement and the coverage rate of the offline dataset. To model the data coverage of a specific offline dataset, we apply t-SNE~\cite{van2008visualizing} to perform dimensionality reduction, then cluster with t-SNE information on all levels of datasets of a given environment, as detailed in Appendix~\ref{apx:coverage_rate}. 

The results illustrate that \methodname\ correlates with the sample coverage rate (See Fig.~\ref{fig:offlinecoverage}, where the left represents the medium-replay level, the middle represents the medium level and the right represents the medium-expert level of the dataset). Middle sample coverage rate yields more significant performance improvement for \methodname, while higher or lower coverage rates lead to less significant performance improvements. This is because lower coverage rates induce a narrow distribution of the offline dataset, resulting in limited information of the offline model. Conversely, higher coverage rates contribute to satisfaction with the offline model, hence further improvement is limited. However, moderate sample coverage rates are common in practical applications.

\section{Related Work}

\textbf{Online RL with assistance of dataset.} Expert-level demonstrations are considered and combined with online data to accelerate the online training process~\cite{hester2018deep, li2023guided, NairDemostrations2018, rajeswaran2018dapg, vecerik2017leveraging}. In contrast, another line of work trains the RL agent from scratch and accelerates the process by sampling both offline and online data~\cite{hester2018deep,song2023hybrid,ball2023efficient,zhou2024offline}. Auxiliary behavioral cloning losses with policy gradients are utilized to guide the updates and accelerate convergence~\cite{kang2018policy,zhu2018reinforcement,zhu2019dexterous}.

\textbf{Offline-to-online RL.} Large task-agnostic datasets are utilized to pre-train a unified offline model and feed for subsequent online RL tasks~\cite{Aytar2018youtube,lowrey2018plan,fujimoto19a,baker2022video,lifshitz2024steve,yuan2024pretraining}. 

A different category of work is to train offline with provided dataset and then fine-tune online in the same environment, which could be further divided as model-based O2O RL and model-free O2O RL. Model-based O2O RL algorithms combined with prioritized sampling scheme~\cite{mao2022moore}, adaptive behavior regularization algorithm MOTO~\cite{rafailov2023moto} or energy-guided diffusion sampling technique~\cite{liu2024energyguided} are proposed to mitigate O2O distribution shift.

For the model-free O2O RL, some offline RL algorithms are directly applied for O2O setting~\cite{nair2020accelerating,kumar2020conservative,kostrikov2022offline}. A series of Q-ensemble based algorithms are proposed, while combined with balanced experience replay~\cite{lee2022offline}, state-dependent balance coefficient~\cite{wang2024train}, uncertainty quantification guidance~\cite{guo2023simple}, uncertainty penalty and smoothness regularization~\cite{wen2023towards} and optimistic exploration~\cite{zhao2024enotoimprovingofflinetoonlinereinforcement}. Recently some work attempts to efficiently explore the environment to accelerate the fine-tuning process: O3F optimistically takes actions with higher expected Q-values~\cite{mark2022fine}, PEX introduces an extra policy to adaptively explore and learn~\cite{zhang2023policy}, OOO framework maintains an exploration policy to collect data and an exploitation policy to train on all data~\cite{mark2024offline} and PTGOOD utilizes planning procedure to explore high-reward areas distant from offline distribution~\cite{mcinroe2024planning}. There are some other independent works: SPOT brings out a density-based regularization term to model the behavior policy~\cite{wu2022supported}, Td3-BC integrates behavioural cloning /constraint that decays over time~\cite{beeson2022improving}, Cal-QL calibrates the learned Q-values at reasonable scale with some reference policy~\cite{nakamoto2023calql}. OLLIE proposes the O2O imitation learning~\cite{yue2024ollie}.

We observe that a cocurrent work WSRL explores to initialize the replay buffer without retaining offline data~\cite{zhou2024efficient}. Nevertheless, \methodname\ proposes a complete and effective solution, whereas WSRL only explores the warm-up of the initial replay buffer.

\section{Conclusion}
\label{sec:conclusion}
This paper proposes a novel paradigm named \methodname\ to eliminate the tedious usage of offline data and leverage the pre-trained offline critic model instead, thereby ensuring
 100\% online sample utilization and better fine-tuning performance.
\methodname\ seamlessly combines online and offline critics with a state-action-conditional coefficient without introducing undesirable or questionable intrinsic rewards. This coefficient estimates the complex distribution of the offline dataset and provides the probability of a given state-action sample.
Theoretical analysis proves the convergence optimality and lower estimation error. Experimental results demonstrate the superiority of \methodname\ over vanilla baselines. However, the performance improvement is limited if the offline dataset distribution is extremely narrow. This limitation could potentially be mitigated by designing specific update strategies for OOD samples, which is an interesting direction for future work. 

\section*{Impact Statement}
This paper presents work whose goal is to advance the field of 
Machine Learning. There are many potential societal consequences 
of our work, none which we feel must be specifically highlighted here.
\bibliographystyle{abbrvnat}
\bibliography{reference.bib}

\newpage
\appendix
\onecolumn
\section{Theoretical analysis}
\label{Apx:theretical_proof}
\subsection{Contraction property}
\label{Apx:contraction_mapping_property}
Our algorithm actually breaks the typical Bellman Equation of the RL algorithm denoted as $Q = \max \limits_{\pi \in \Pi} \bellman^\pi = \max \limits_{\pi \in \Pi} (r_\pi + \gamma P_\pi Q)$. Instead we promote Eq. (\ref{eqn:efom_initial_equation}). In order to prove the convergence of the updating equation, we introduce the contraction mapping theorem which is widely used to prove the convergence optimality of RL algorithm.

\begin{theorem}[Contraction mapping theorem]
\label{thm:contraction mapping theorem}
For an equation that has the form of $x = f(x)$ where $x$ and $f(x)$ are real vectors, if $f$ is a contraction mapping which means that $\Vert f(x_1) - f(x_2)\Vert \leq  \gamma \Vert x_1 - x_2\Vert (0<\gamma <1)$, then the following properties hold.

Existence: There exists a fixed point $x^*$ that satisfies $f(x^*) = x^*$.

Uniqueness: The fixed point $x^*$ is unique.

Algorithm: Given any initial state $x_0$, consider the iterative process: $x_{k+1} = f(x_k)$, where $k=0, 1, 2, ...$. Then $x_k$ convergences to $x^*$ as $k \to \infty$ at an exponential convergence rate.

\end{theorem}

We just need to prove that this equation satisfies the contraction property of theorem ~\ref{thm:contraction mapping theorem} and naturally we can ensure the convergence of the algorithm.

Take the right hand of Eq. (\eqref{eqn:efom_initial_equation}) as function $f(Q)$ and consider any two vectors $Q_1, Q_2 \in \mathbb{R}^{\mathcal{S}}$, and suppose that $\pi_1^* \doteq \mathop{\arg\max}\limits_\pi (r_\pi + \gamma P_\pi Q_1)$, $\pi_2^* \doteq \mathop{\arg\max}\limits_\pi (r_\pi + \gamma P_\pi Q_2)$. Then,
\begin{equation}
    \begin{split}
        f(Q_1) =& \max \limits_\pi \left[(1 -p(s, a)) \bellman ^\pi Q_1 + p(s, a) \bellman ^\pi Q^{off})\right]\\
        =& (1 - p(s, a)) \bellman ^{\pi_1^*} Q_1 + p(s, a) \bellman ^{\pi_1^*} Q^{off}\\
        \geq & (1 - p(s, a)) \bellman ^{\pi_2^*} Q_1 + p(s, a) \bellman ^{\pi_2^*} Q^{off},
    \end{split}
\end{equation}
and similarly: 
\begin{equation}
    \begin{split}
        f(Q_2) \geq (1 - p(s, a)) \bellman ^{\pi_1^*} Q_2 + p(s, a) \bellman ^{\pi_1^*} Q^{off}.
    \end{split}
\end{equation}

To simplify the derivation process, we use $p^{\pi}$ to represent $p(s, \pi(a|s))$ considering that values of $p$ function class are determined by the policy $\pi$ of any given state. As a result,
\begin{equation}
    \begin{split}
        &f(Q_1) - f(Q_2) \\
        =& (1 - p^{\pi_1^*}) \bellman ^{\pi_1^*} Q_1 + p^{\pi_1^*} \bellman ^{\pi_1^*} Q^{off}- \left[ (1 - p^{\pi_2^*}) \bellman ^{\pi_2^*} Q_2 + p^{\pi_2^*} \bellman ^{\pi_2^*} Q^{off} \right]\\
        \leq& (1 - p^{\pi_1^*}) \bellman ^{\pi_1^*} Q_1 + p^{\pi_1^*} \bellman ^{\pi_1^*} Q^{off} - \left[(1 - p^{\pi_1^*}) \bellman ^{\pi_1^*} Q_2 + p^{\pi_1^*} \bellman ^{\pi_1^*} Q^{off}\right] \\
        =& (1 - p^{\pi_1^*}) (\bellman ^{\pi_1^*} Q_1 - \bellman ^{\pi_1^*} Q_2)\\
         =&\gamma (1 - p^{\pi_1^*}) P^{\pi_1^*}(Q_1 - Q_2)\\
        \leq& \gamma P^{\pi_1^*}(Q_1 - Q_2).
    \end{split}
\end{equation}

We can see that the result reduces to that of the normal Bellman equation and therefore, the following derivation is omitted. As a result, we get,
\begin{equation}
    \begin{split}
         \Vert f(Q_1) - f(Q_2)\Vert_\infty \leq  \gamma \Vert Q_1 - Q_2\Vert_\infty,
    \end{split}
\end{equation}
which concludes the proof of the contraction property of $f(Q)$.

\subsection{Convergence optimality}
\label{Apx:ConvergenceAnalysis}
We consider a tabular setting for simplicity. We first write down the iterative form of Eq. (\ref{eqn:efom_final_equation}) as below:

\noindent if $s=s_k, a = a_k$,
\begin{equation}
    \label{eqnapx:theoretical_iterative_expression}
    \begin{split}
        Q_{k+1} (s, a)=&Q_{k} (s, a)-\alpha_k (s, a) \bigl[Q_k(s, a) -\left( r_{k+1} + \gamma Q_k (s_{k + 1}, a_{k + 1})\right) \bigr]\\
        &- \alpha_k (s, a)\gamma p(s, a) \bigl( Q^{off}(s_{k + 1}, a_{k + 1}) - Q_k(s_{k + 1}, a_{k + 1})\bigr).
    \end{split}
\end{equation}
else,
\begin{equation}
    \label{eqnapx:theoretical_nonupdate_expression}
    \begin{split}
        Q_{k+1} (s, a) = Q_{k} (s, a).
    \end{split}
\end{equation}

The error of estimation is defined as:
\begin{equation}
    \begin{split}
        \Delta_k (s, a) \doteq Q_k(s, a) - Q(s, \pi).
    \end{split}
\end{equation}
where $Q_\pi(s, a)$  is the state action value s under policy $\pi$. Deducting $Q_\pi(s, a)$ from both sides of~\ref{eqn:theoretical_iterative_expression} gets:
\begin{equation}
    \label{eqnapx:theoretical_delta}
    \begin{split}
        \Delta_{k+1} (s, a) = &\left( 1 - \alpha_k(s, a)\right) \Delta_k(s, a)+ \alpha_k(s, a)\eta_k(s, a), \quad s=s_k, a=a_k.
    \end{split}
\end{equation}
where 
\begin{equation}
    \label{eqnapx:theoretical_eta}
    \begin{split}
&\eta_k(s, a) \\
=&\bigl[r_{k+1} + \gamma Q_k (s_{k + 1}, a_{k + 1}) - Q^\pi(s, a)\bigr] + \gamma p(s, a) \bigl[ Q^{off}(s_{k + 1}, a_{k + 1}) - Q_k(s_{k + 1}, a_{k + 1})\bigr]\\
=&\bigl[r_{k+1} + \gamma Q_k (s_{k + 1}, a_{k + 1}) - Q^\pi(s, a)\bigr] +\\
&\gamma p(s, a) \Bigl\{ \left[Q^{off}(s_{k + 1}, a_{k + 1}) -Q^\pi(s_{k + 1}, a_{k + 1})\right] + \bigl[Q^\pi(s_{k + 1}, a_{k + 1}) - Q_k(s_{k + 1}, a_{k + 1})\bigr]\Bigr\}\\
        =&\underbrace{ \bigl[r_{k+1} + \gamma Q_k (s_{k + 1}, a_{k + 1}) - Q^\pi(s, a)\bigr]}_{\Gamma_1}- \underbrace{ \gamma p(s, a) \bigl[ Q_k(s_{k + 1}, a_{k + 1}) - Q^\pi(s_{k + 1}, a_{k + 1})\bigr]}_{\Gamma_2}\\
        &+\underbrace{ \gamma p(s, a) \bigl[ Q^{off}(s_{k + 1}, a_{k + 1}) - Q^\pi(s_{k + 1}, a_{k + 1})\bigr]}_{\Gamma_3}\\
        =&\Gamma_1 - \Gamma_2 + \Gamma_3.
    \end{split}
\end{equation}

Similarly, deducting $Q^\pi(s, a)$ from both side of Eq. (\ref{eqnapx:theoretical_nonupdate_expression}) gets:
\begin{equation}
    \nonumber
    \begin{split}
        \Delta_{k+1} (s, a) =& \left( 1 - \alpha_k(s, a)\right) \Delta_k(s, a)+ \alpha_k(s, a)\eta_k(s, a), \quad s\neq s_k\ or\ a \neq a_k.
    \end{split}
\end{equation}
this expression is the same as~\ref{eqnapx:theoretical_delta} except that $\alpha_k(s, a)$ and $\eta_k(s, a)$ is zero. Therefore we observe the following unified expression:
\begin{equation}
    \nonumber
    \begin{split}
        \Delta_{k+1} (s, a) = \left( 1 - \alpha_k(s, a)\right) \Delta_k(s, a) + \alpha_k(s, a)\eta_k(s, a).
    \end{split}
\end{equation}

To further analyze the convergence property, we introduce Dvoretzky's theorem~\cite{convergence1994}:
\begin{theorem}[Dvoretzky's Throrem]
\label{thm:Dvoretzky's Throrem}
Consider a finite set $\mathcal{S}$ of real numbers. For the stochastic process:
\begin{equation}
    \nonumber
    \begin{split}
        \Delta_{k+1} (s) = \left( 1 - \alpha_k(s)\right) \Delta_k(s) + \beta_k(s)\eta_k(s).
    \end{split}
\end{equation}
it holds that $\Delta_k(s)$ convergences to zero almost surely for every $s \in \mathcal{S}$ if the following conditions are satisfied for $s \in \mathcal{S}$:
\begin{itemize}
\item[(a)] $\sum_k \alpha_k(s) = \infty,  \sum_k \alpha_{k}^{2}(s) < \infty,$ $\sum\limits_{k} \beta_{k}^{2}(s) < \infty$, $\mathbb{E}[\beta_k(s)|\mathcal{H}_k] \leq \mathbb{E}[\alpha_k(s)|\mathcal{H}_k]$ uniformly almost surely;
\item[(b)] $\big\Vert{\mathbb{E}[\eta_k(s)|\mathcal{H}_k]}\big\Vert_\infty \leq \gamma \big\Vert \Delta_k \big\Vert_\infty$, with $\gamma \in (0, 1)$;
\item[(c)] $var \left[ \eta_k(s) | \mathcal{H}_k \right] \leq C\left( 1 + \big\Vert \Delta_k(s) \big\Vert_\infty \right)^2$, with $C$ a constant.
\end{itemize}
Here, $\mathcal{H}_k=\{ \Delta_{k}, \Delta_{k-1}, \cdots, \eta_{k-1}, \cdots, \alpha_{k-1},\cdots, \beta_{k-1},$ $\cdots\}$ denotes the historical information. The term $\Vert \cdot \Vert_\infty$ represents the maximum norm.
\end{theorem}

To prove \methodname\ is well-converged, we just need to validate that the three conditions are satisfied. Nothing changes in our algorithm compared to normal RL algorithms when considering the first condition so it is naturally satisfied. Please refer to~\cite{convergence1994} for detailed proof. For the second condition, due to the Markovian property, $\eta_t(s, a)$ does not depend on the historical information and is only dependent on $s$ and $a$. Then, we get $\mathbb{E}[\eta_k(s, a)|\mathcal{H}_k] = \mathbb{E}[\eta_k(s, a)]$. 

Specifically, for $s=s_t, a = a_t$, we have:
\begin{equation}
    \nonumber
    \begin{split}
        \mathbb{E}[\eta_k(s, a)] =\mathbb{E}[\eta_k(s_k, a_k)]=\mathbb{E}[\Gamma_1] - \mathbb{E}[\Gamma_2] + \mathbb{E}[\Gamma_3].
    \end{split}
\end{equation}

For the first  term,
\begin{equation}
    \nonumber
    \begin{split}
        \mathbb{E}[\Gamma_1] =& \mathbb{E}\bigl[r_{k+1} + \gamma Q_k (s_{k + 1}, a_{k + 1}) - Q^\pi(s_k, a_k)\left|s_k, a_k\right.\bigr]\\
        =& \mathbb{E}\bigl[r_{k+1} + \gamma Q_k (s_{k + 1}, a_{k + 1})\left|s_k, a_k\right.\bigr] - Q^\pi(s_k, a_k).
    \end{split}
\end{equation}
Since $Q_\pi(s_k, a_k) = \mathbb{E}\left[r_{k+1} + \gamma Q^\pi (s_{k + 1}, a_{k + 1})\left|s_k, a_k\right.\right]$, the above equation indicates that,
\begin{equation}
    \nonumber
    \begin{split}
        \mathbb{E}[\Gamma_1] =& \gamma \mathbb{E}\bigl[ Q_k(s_{k + 1}, a_{k + 1}) - Q^\pi(s_{k + 1}, a_{k + 1})\left|s_k, a_k\right.\bigr].
    \end{split}
\end{equation}

For the second term,
\begin{equation}
    \nonumber
    \begin{split}
        \mathbb{E}[\Gamma_2] =& \gamma p(s_k, a_k) \mathbb{E}\bigl[ Q_k(s_{k + 1}, a_{k + 1})-Q^\pi(s_{k + 1}, a_{k + 1})\left|s_k, a_k\right.\bigr].
    \end{split}
\end{equation}

Combining these two terms gets:
\begin{equation}
    \nonumber
    \begin{split}
        \mathbb{E}[\Gamma_1] - \mathbb{E}[\Gamma_2]=& \gamma \left(1-p(s_k, a_k)\right) \mathbb{E}\bigl[ Q_k(s_{k + 1}, a_{k + 1}) - Q^\pi(s_{k + 1}, a_{k + 1})\left|s_k, a_k\right.\bigr].
    \end{split}
\end{equation}

Then,
\begin{equation}
    \nonumber
    \begin{split}
         &\big\Vert\mathbb{E}[\Gamma_1] - \mathbb{E}[\Gamma_2]\big\Vert_\infty\\
         =&\gamma \left(1-p(s_k, a_k)\right) \Big\Vert\sum \limits_{a^\prime \in \mathcal{A}} \sum \limits_{s^\prime \in \mathcal{S}}  t\left(s^\prime, a^\prime \left| s_k, a_k \right.\right) \bigl| Q_k(s^\prime, a^\prime)- Q^\pi(s^\prime, a^\prime) \bigr|\Big\Vert_\infty\\
         =&\gamma  \left(1-p(s_k, a_k)\right) \max \limits_{s^\prime \in \mathcal{S}, a^\prime \in \mathcal{A}} \bigg\{ \sum \limits_{a^\prime \in \mathcal{A}} \sum \limits_{s^\prime \in \mathcal{S}} t\left(s^\prime, a^\prime \left| s_k, a_k \right.\right) \bigl| Q_k(s^\prime, a^\prime)- Q^\pi(s^\prime, a^\prime) \bigr|\bigg\}\\
         \leq& \gamma \left(1-p(s_k, a_k)\right)\sum \limits_{a^\prime \in \mathcal{A}} \sum \limits_{s^\prime \in \mathcal{S}}  t\left(s^\prime, a^\prime \left| s_k, a_k \right.\right) \max \limits_{s^\prime \in \mathcal{S}, a^\prime \in \mathcal{A}}\Bigl[ \bigl| Q_k(s^\prime, a^\prime)- Q^\pi(s^\prime, a^\prime) \bigr|\Bigr]\\
        =&\gamma \left(1-p(s_k, a_k)\right)\max \limits_{s^\prime \in \mathcal{S}, a^\prime \in \mathcal{A}} \Bigl[\bigl| Q_k(s^\prime, a^\prime)- Q^\pi(s^\prime, a^\prime)\bigr| \Bigr]\\
        =&\gamma \left(1-p(s_k, a_k)\right) \big\Vert \Delta_k(s, a) \big\Vert_\infty.
    \end{split}
\end{equation}

For the third term, to simplify the derivation, we mildly abuse the notation of $s^\prime, a^\prime$ to represent $s_{k + 1}, a_{k + 1}$, 
\begin{equation}
    \nonumber
    \begin{split}
         \mathbb{E}[\Gamma_3]=&\mathbb{E}\left[\gamma p(s_k, a_k) \left( Q^{off}(s_{k + 1}, a_{k + 1}) - Q^\pi(s_{k + 1}, a_{k + 1})\right)\right]\\
         =&\gamma \sum \limits_{a^\prime \in \mathcal{A}} \sum \limits_{s^\prime \in \mathcal{S}}  t\left(s^\prime, a^\prime \left| s_k, a_k \right.\right) p(s_k, a_k) \left| Q^{off}(s^\prime, a^\prime)- Q^\pi(s^\prime, a^\prime) \right|.
    \end{split}
\end{equation}

It follows that:
\begin{equation}
    \nonumber
    \begin{split}
         &\big\Vert\mathbb{E}[\Gamma_3]\big\Vert_\infty\\
         =&\Big\Vert\gamma \sum \limits_{a^\prime \in \mathcal{A}} \sum \limits_{s^\prime \in \mathcal{S}}  t\left(s^\prime, a^\prime \left| s_k, a_k \right.\right) p(s_k, a_k) \bigl| Q^{off}(s^\prime, a^\prime)- Q^\pi(s^\prime, a^\prime) \bigr|\Big\Vert_\infty\\
         =&\gamma  p(s_k, a_k) \max \limits_{s^\prime \in \mathcal{S}, a^\prime \in \mathcal{A}}\bigg\{ \sum \limits_{a^\prime \in \mathcal{A}} \sum \limits_{s^\prime \in \mathcal{S}} t\left(s^\prime, a^\prime \left| s_k, a_k \right.\right) \bigl| Q^{off}(s^\prime, a^\prime)- Q^\pi(s^\prime, a^\prime) \bigr|\bigg\}\\
         \leq& \gamma p(s_k, a_k)\sum \limits_{a^\prime \in \mathcal{A}} \sum \limits_{s^\prime \in \mathcal{S}}  t\left(s^\prime, a^\prime \left| s_k, a_k \right.\right) \max \limits_{s^\prime \in \mathcal{S}, a^\prime \in \mathcal{A}}\Bigl[ \bigl| Q^{off}(s^\prime, a^\prime)- Q^\pi(s^\prime, a^\prime) \bigr|\Bigr]\\
        =&\gamma p(s_k, a_k)\max \limits_{s^\prime \in \mathcal{S}, a^\prime \in \mathcal{A}} \Bigl[ \bigl| Q^{off}(s^\prime, a^\prime)- Q^\pi(s^\prime, a^\prime)\bigr|\Bigr].
    \end{split}
\end{equation}

If the sample $(s^\prime, a^\prime)$ is in the distribution of offline dataset, We notice that the probability $(s^\prime, a^\prime)$ is significant and the $Q_k(s^\prime, a^\prime)$ is a good estimation of the optimal value $Q_\pi(s^\prime, a^\prime)$ and the specific form of TD error depends on the offline algorithm, and we can uniformly formulate this by:
\begin{equation}
    \nonumber
    \begin{split}
        \big\Vert\mathbb{E}[\Gamma_3]\big\Vert_\infty &= \gamma \gamma_{\mathcal{F}}p(s_k, a_k) \max \limits_{s^\prime, a^\prime \in \mathcal{D}} \Bigl[\bigl| Q^{off}(s^\prime, a^\prime)- Q^\pi(s^\prime, a^\prime)\bigr| \Bigr]\\
        &=\gamma \gamma_{\mathcal{F}}p(s_k, a_k) \big\Vert \Delta_{N}(s, a) \big\Vert_\infty.
    \end{split}
\end{equation}
where $\mathcal{F}$ denotes the function class of offline algorithm, $0<\gamma_{\mathcal{F}}<1$ denotes the convergence coefficient of offline algorithm class $\mathcal{F}$ and $N$ denotes the iterative number of offline pre-training.

But while $(s^\prime, a^\prime)$ falls out of the distribution of offline dataset, the probability $p(s^\prime, a^\prime)$ is trivial with an upper bound constrained to a diminutive number $\xi_{OOD}$, denoted as $p(s^\prime, a^\prime)<\xi_{OOD}(s^\prime, a^\prime)$, and we know little about the $\left| Q_\pi(s^\prime, a^\prime)- Q_k(s^\prime, a^\prime) \right|$ but it is inherently restricted by the maximum reward $R_{max}$. Then this term is limited by $2\gamma\xi_{OOD}(s^\prime, a^\prime)R_{max}$ and we cut the probability $p(s^\prime, a^\prime)$ to zero in practice. Combining the above two cases gets the following upper limit:
\begin{equation}
    \nonumber
    \begin{split}
         \big\Vert\mathbb{E}[\Gamma_3]\big\Vert_\infty\leq& \max \Bigl\{\gamma \gamma_{\mathcal{F}}p(s_k, a_k) \big\Vert \Delta_{N}(s, a) \big\Vert_\infty, 2\gamma\xi_{OOD}(s^\prime, a^\prime)R_{max}\Bigr\}\\
        =&\gamma \gamma_{\mathcal{F}}p(s_k, a_k) \big\Vert \Delta_{N}(s, a) \big\Vert_\infty.
    \end{split}
\end{equation}
 
Therefore,
\begin{equation}
    \nonumber
    \begin{split}
        \big\Vert\mathbb{E}[\eta_k(s, a)]\big\Vert_\infty =&\big\Vert\mathbb{E}[\Gamma_1] - \mathbb{E}[\Gamma_2] + \mathbb{E}[\Gamma_3]\big\Vert_\infty\\
        \leq& \gamma \left(1-p(s_k, a_k)\right) \big\Vert \Delta_k(s, a) \big\Vert_\infty + \gamma \gamma_{\mathcal{F}}p(s_k, a_k) \big\Vert \Delta_{N}(s, a) \big\Vert_\infty.
    \end{split}
\end{equation}

Because $N$ is big enough that $\big\Vert \Delta_{N}(s, a) \big\Vert_\infty$ is a high-order small quantity compared to $\big\Vert \Delta_k(s, a) \big\Vert_\infty$ and can be written as $\mathcal{O}(\big\Vert \Delta_k(s, a) \big\Vert_\infty)$. Therefore,
\begin{equation}
    \label{apxeqn:contraction_factor}
    \begin{split}
        \big\Vert\mathbb{E}[\eta_k(s, a)]\big\Vert_\infty \leq& \gamma \left(1-p(s_k, a_k)\right) \big\Vert \Delta_k(s, a) \big\Vert_\infty + \gamma \gamma_{\mathcal{F}}p(s_k, a_k)\mathcal{O}(\big\Vert \Delta_k(s, a) \big\Vert_\infty.
    \end{split}
\end{equation}
where $0<\gamma \left(1-p(s_k, a_k)\right)<1$ and the second condition is satisfied. Finally, regarding the third condition, we have when $s = s_k, a + a_k$,
\begin{equation}
    \nonumber
    \begin{split}
        &var \bigl[ \eta_k(s) | \mathcal{H}_k \bigr] \\
        =& var\Bigl\{ \bigl[r_{k+1} + \gamma Q_k (s_{k + 1}, a_{k + 1}) - Q^\pi(s, a)\bigr] + \gamma p(s, a) \bigl[ Q^{off}(s_{k + 1}, a_{k + 1}) - Q_k(s_{k + 1}, a_{k + 1})\bigr]\Bigr\}.
    \end{split}
\end{equation}
and $var \left[ \eta_k(s) | \mathcal{H}_k \right]=0$ for $s\neq s_k$ or $a\neq a_k$.

Since $r_{k+1}$ and $\mathbb{E} \left[ \eta_k(s) | \mathcal{H}_k \right]$ are both bounded, the third condition can be proven easily. And Therefore \methodname\ is well converged.

\subsection{Convergence Speed}
\label{Apx:convergence_speed}
In this section, we give a more detailed analysis of the convergence speed of \methodname. For vanilla RL algorithms, the contraction coefficient $\gamma$ represents the convergence speed because it controls the contraction speed of Q-iteration. For \methodname, we give a rough derivation in Appendix~\ref{Apx:contraction_mapping_property} that \methodname\ possesses a smaller contraction coefficient. But how small could that be? We actually have already derived the specific form of contraction factor in Appendix~\ref{Apx:ConvergenceAnalysis}, as specified in Eq. (\ref{apxeqn:contraction_factor}). However, Eq. (\ref{apxeqn:contraction_factor}) just covers the in-distribution situation of the contraction coefficient. As for the OOD situation, the contraction coefficient share the same coefficient as the normal Bellman equation. To sum up, we write the whole the contraction coefficient as below:
\begin{align}
    &\gamma_{o} \leq \left\{
    \begin{aligned}
    \label{eqnapx:TD_error_term}
    &\begin{aligned}
        &\left(1-p(s_k, a_k)\right)\gamma + \gamma \gamma_{\mathcal{F}}p(s_k, a_k)\frac{\big\Vert \Delta_{off}(s, a) \big\Vert_\infty}{\big\Vert\Delta_k(s, a) \big\Vert_\infty},
    \end{aligned}
     &p(s_k, a_k)\geq p_m\\
    &\gamma, &p(s_k, a_k)<p_m
    \end{aligned}
    \right.
\end{align}
where $0<\gamma_{\mathcal{F}}<1$ denotes the convergence coefficient of offline algorithm class $\mathcal{F}$ and $\big\Vert \Delta_{off}(s, a) \big\Vert_\infty$ denotes the offline suboptimality bound $\left[V^*(s) - V^{\pi_{off}}(s)\right]$ and $\big\Vert\Delta_k(s, a) \big\Vert_\infty$ denotes the suboptimality bound of the k-th iteration of online fine-tuning.

The upper equation holds for in-distribution samples, which are well mastered by the offline model. Therefore, the offline suboptimality bound is substantially tighter compared to the online bound. This illustrates that the offline model guidance significantly accelerates the online fine-tuning process by providing accurate estimations for in-distribution samples.

\section{State-action-conditional coefficient}
\label{apx:state_action_cofficient}
\subsection{C-VAE details}
\subsubsection{Posterior collapse situation}
\label{apx:vae_posterior_collapse}
\begin{figure}[t] 
\centering
{\includegraphics[clip,width=0.75\linewidth]{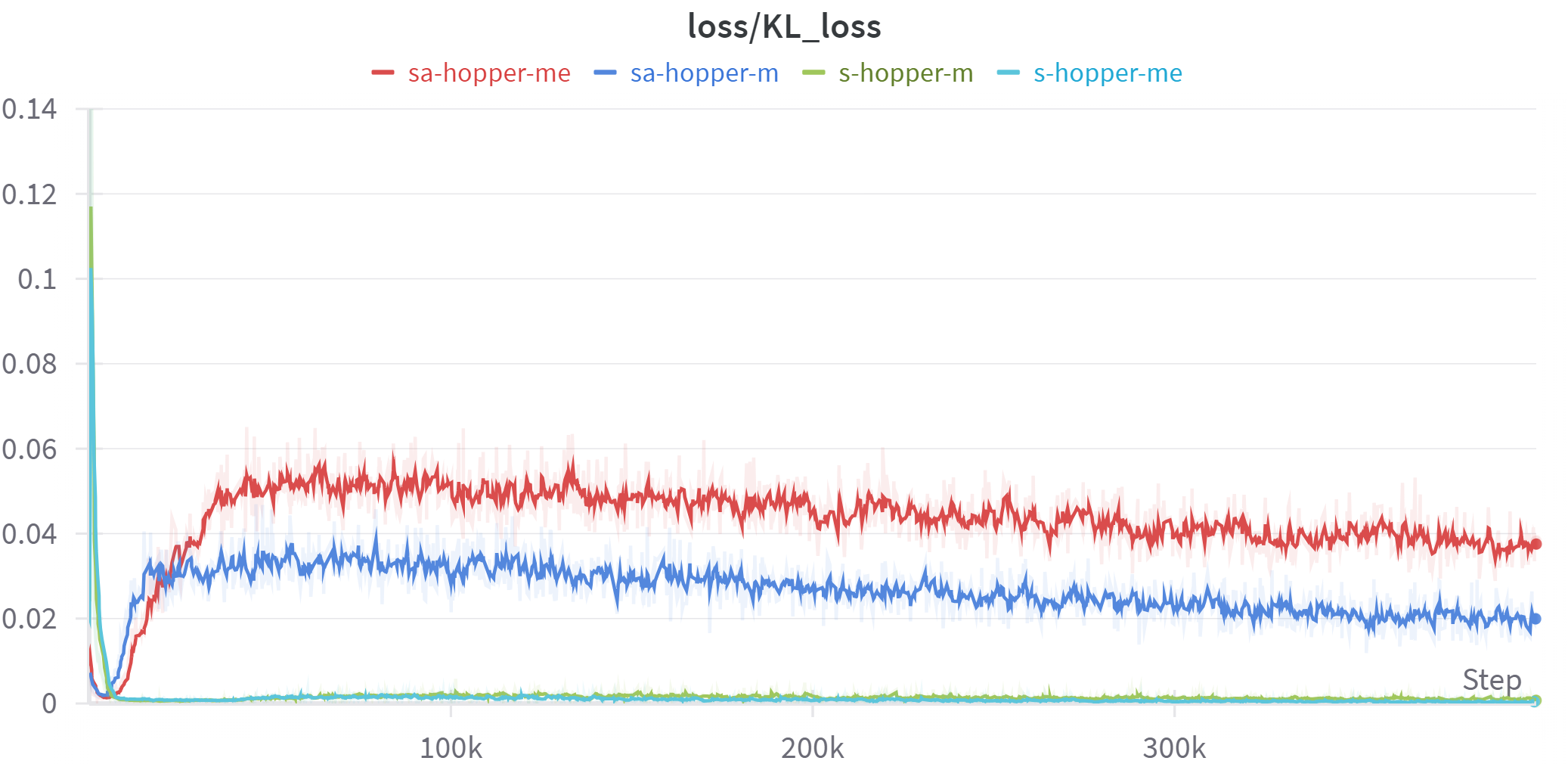}} 
\caption{\label{fig:KL_Loss} \footnotesize{\textbf{Illustration of the posterior collapse of C-VAE structure} The blue curve represents the normal KL loss term while the green term represents the posterior collapse situation.}}
\end{figure}

As stated in Section~\ref{sec:empirical-method}, C-VAE may meet with posterior collapse, as shown in Fig.~\ref{fig:KL_Loss}. Fig.~\ref{fig:KL_Loss} illustrates the KL loss values with posterior collapse (with ``s-hopper-m" and ``s-hopper-me" legend) and without posterior collapse (with ``sa-hopper-m" and ``sa-hopper-me" legend), representing the distribution error between the output distribution of encoder and standard normal distribution. It can be observed that situations with posterior collapse possess much lower loss term (approximately by four orders of magnitude). Though the loss is lower for the posterior collapse situation, the output of all state-action samples are 0 and 1 for mean and standard respectively. Therefore the encoder totally fails to function.

\subsubsection{VAE implementations}
\label{apx:vae_implementations}
For the C-VAE module, we employ the same VAE structure as Xu~\cite{xu2022constraints} except that we change the input to (state, action) and the output to next state. Furthermore, we adopt the KL-annealing technique in the hopper environments where we do not introduce the KL loss initially by manually setting it to zero and slowly increasing the KL loss weight with time. KL-annealing could result in more abundant representations of the encoder and is less likely to introduce posterior collapse. We also simplify the decoder of the C-VAE module in hopper and Walker2d environments to avoid posterior collapse. Notably, avoid normalizing the states and the actions because the normalized states are highly likely to result in the posterior collapse. In terms of experimental experience, the algorithm performs best when the KL loss converges to around 0.03. The information of the next state is supplemented in the training phase to better model the offline distribution and statistical techniques are combined with neural networks to obtain more reasonable probability estimation. 

\subsubsection{Practical VAE distribution}
\label{apx:practical_VAE_distribution}

\begin{figure}[t] 
\centering
{\includegraphics[clip,width=0.65\linewidth]{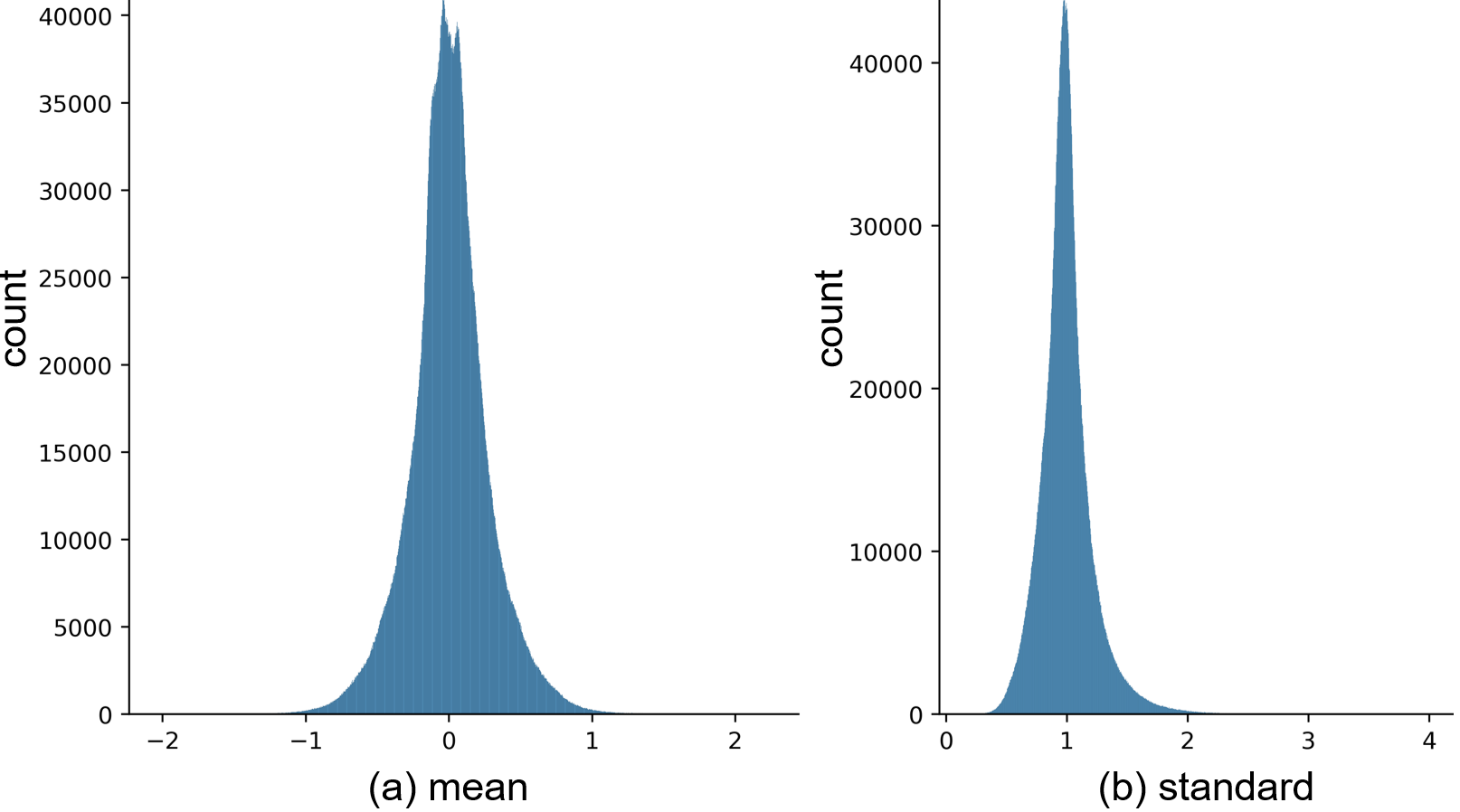}}
\caption{\label{fig:mean_standard}\footnotesize{\textbf{Statistical results of the output from the C-VAE model, including (a) the mean values and (b) the standard values}}}
\end{figure}

In practice implementation, the offline data is the input to C-VAE model, and the statistical result of output from the C-VAE model, including mean and standard, are shown in the Fig.~\ref{fig:mean_standard}.

\subsubsection{Adaptive VAE Coefficient}
\label{apx:adaptive_vae_coefficient}
By adding a weight variable to the replay buffer, we store the offline critic value and $p^{off}$ for each sample. At fixed intervals, we first collect data from the current period (all data from the previous to the current interval) and then filter out OOD samples, whose $p^{off}$ is equal to 0. The challenge lies in identifying the mastered OOD samples. Since the model is lack of awareness of OOD samples, estimated Q-values tend to introduce significant errors and, consequently, large loss terms during training. This can destroy the model and reduce algorithm performance. Therefore, the magnitude of the error between the estimated critic during online fine-tuning and the true Q-values can be used as a measure of how well the OOD samples have been mastered. However, in practice, the ground truth of the Q-values is unavailable, so the exact error can not be obtained. To address this, we adopted several potential approaches to estimate the error, which will be detailed later. After obtaining error estimates, we select samples with minimal errors as mastered OOD samples. These samples are then used to continue training the VAE model to help it learn these samples.

In the actual implementation, the error estimation methods replace the true Q-values with sampling (as referenced in~\cite{kumar2020conservative}) and use the practical Bellman operator, where the target Q-value serves as an estimate. We found that the results of these two methods are similar, with over 70\% overlap in the filtered OOD samples, and both lead to comparable improvements in algorithm performance (with the sampling-based method performing slightly better). Considering the trade-off between the computation overhead and algorithm performance, we choose to use the Bellman operator. Additionally, we set the update interval to 10,000 steps. Since this interval exceeds the target Q network update interval (typically set to 1,000), we consistently refer to the target-Q network at the beginning of the period to ensure fairness. Given that the target network is already saved at this point, no additional computation overhead is introduced. Minimal error is defined as the smallest 10\% of errors among the filtered samples. 

\subsection{Embedding network}
\label{apx:embedding_network}
The embedding network structure is inspired by the work of Badia~\cite{badia2019never}, where this network is introduced to extract valuable information from two successive states and predict the probability of all the actions. Thinking that we need to model the offline distribution of the state-action pairs, it is inspiring to construct a representation from two structurally consistent and logically connected states to predict the actions with different structures. It is worth noting that Badia only discusses the discrete environments, implying that the actions are represented with one-hot encoding. However, the pipeline conflicts with our setting that the state and action are continuous, we modify the embedding network by computing the loss with the real action values instead of the action representations. In practice, we train a Siamese network to get the controllable states and utilize the controllable states representation to predict the offline probability of each action, termed as the conditional likelihood $p_o(a|s_k, s_{k+1})$, where $h$ is a classifier network followed by a ReLU layer (different from the softmax layer from Badia because the range of actions is not (0, 1)). Therefore, we could only get the loss term instead of the probability because we substituted the softmax layer for ReLU layer. So we also calculate the output of the offline dataset and model the data with a normal distribution, like that in the C-VAE.

\section{Algorithm Implementation}
\label{Apx:algorithm_implementation}
\subsection{\methodname\ implementation on baselines}
\label{Apx:implementations_on_baselines}
To illustrate the whole procedure of \methodname, we first represent the pseudo-code of \methodname\ implemented based on AWAC~\cite{nair2020accelerating} below:
    \begin{algorithm}[H]
        \label{alg:waac_omda_pseudo}
        \caption{Offline-to-Online Reinforcement Learning via State-Action-Conditional Offline Model Guidance (Implemented on AWAC)}
        \begin{algorithmic}[1]
       	\REQUIRE offline Q-network $Q_\phi^{off}$, policy $\pi_\theta^{off}$ and VAE
       	\STATE $\pi_\theta \gets \pi_\theta^{off}$ , $Q_\phi \gets Q_\phi^{off}$ 
       	\STATE Initialize the replay buffer $D$ with $N$ samples collected by $Q_\phi$
       	\FOR{iteration $i=1, 2, ...$}
                \FOR{every environment step}
                    \STATE $a_t \sim \pi_\phi(s_t|s_t)$
                    \STATE $s_{t+1}, d_t \sim p_(s_{t+1}|s_t, a_t)$
                    \STATE insert $(s_t, a_t,r(s_t, a_t), s_{t+1}, d_t)$ into D
                \ENDFOR
                \FOR{every update step}
                    \STATE get $Q_{target}$ and $Q_{target}^{off}$ according to AWAC
                    \STATE get $p_{om}$ with C-VAE according to Section~\ref{VAE_section}
                    \STATE $Q_{target} \gets (1-p_{om})Q_{target} + p_{om}Q_{target}^{off}$
                    \STATE Update $\phi$ according to Eq. 9 in~\cite{nair2020accelerating} with $Q_{target}$
                    \STATE Update $\theta$ according to Eq. 13 in~\cite{nair2020accelerating}
                    \IF{This step is the interval to update coefficient}
                    \STATE Filter mastered OOD samples
                    \STATE Train the VAE model with collected samples
                    \STATE Update the VAE model and the offline critic
                    \ENDIF
                \ENDFOR
           \ENDFOR 
    \end{algorithmic}
    \end{algorithm}

 
    

Only some minimal adjustments are needed to implement \methodname\ on AWAC, and IQL~\cite{kostrikov2022offline} as well. We just need to maintain a much smaller replay buffer filled with online samples and insert and sample from this ``online replay buffer". Before conducting normal gradient update step, we need to calculate the mixed $Q_{target}$ according to Eq. (\ref{eqn:efom_final_equation}). As for IQL, we freeze and query the offline pre-trained value function instead because IQL separately trains a value function to serve as the target information. Other implementations are similar to AWAC and are omitted.

As for CQL and Cal\_QL, these two algorithms share a similar implementation procedure and align with AWAC when calculating the $Q_{target}$. However, CQL adds one extra penalty term to minimize the expected Q-value based on a distribution $\mu(a|s)$, formulated as $\mathbb{E}_{s\sim \mathcal D, a\sim \mu(a|s)}\left[ Q(s, a)\right]$. This term is separated from the standard Bellman equation and serves an important role in making sure the learned Q-function is lower-bounded. However, this term is unrestricted in our paradigm and may cause algorithm divergence. So we add an offline version of the term still weighted by the state-action-conditional coefficient. This slightly avoids our setting but is reasonable that this setting shares the consistent updating direction with the Bellman equation error term.

We implement all the algorithms based on the benchmark CORL~\cite{tarasov2024corl}, whose open source code is available at~\url{https://github.com/tinkoff-ai/CORL} and the license is Apache License 2.0 with detail in the GitHub link. Our code is attached in the supplementary material.

\subsection{Implementation Details}
\label{Apx:implementation_details}
In practice, we strictly adopt the CORL setting to train the offline model and the vanilla fine-tuning training, including the training process and hyperparameters. As for \methodname\ training, for mujoco environments (halfcheetah, hopper, walker2d), \methodname\ algorithms share the same set of hyperparameters with the fine-tuning process to illustrate fairness. In the antmaze environment, we slightly reduce the weight of the Q-value maximization term to highlight the impact of \methodname\ for algorithms CQL and Cal\_QL, from 5 to 2. For the threshold $p^{vae}_{m}$, we adopt the value of 0.6 in most environments (including HalfCheetah, Hopper, Walker2d and Adroit) which seems large but only a small portion satisfies the condition. For the antmaze environment, we take 0.7 for CQL and 0.6 for the others. We found that in our setting, reducing the size of the replay buffer allows for more efficient utilization of samples, thereby improving the algorithm's performance. Specifically, we set the buffer size to be 50,000 for Anrmaze environment and 20,000 for the other environments. We initialize the replay buffer with 2000 samples utilizing the offline model (2000 is the normal length of an episode in most environments). The details of C-VAE have been stated in Appendix~\ref{apx:state_action_cofficient}.

\subsection{Datasets}
\label{apx:dataset_description}
D4RL (Datasets for Deep Data-Driven Reinforcement Learning)~\cite{fu2020d4rl} is a standard benchmark including a variety of environments. \methodname\ is tested across four environments within D4RL: HalfCheetah, hopper, walker2d and antmaze.
\begin{enumerate}
    \item[1] \textbf{HalfCheetah}: The halfcheetah environments simulates a two-legged robot similar to a cheetah, but only with the lower half of the cheetah. The goal is to navigate and move forward by coordinating the movements of its two legs. It is a challenging environment due to the complex dynamics of the motivation.
    \item[2] \textbf{Hopper}: In the Hopper environment, the agent is required to control a one-legged hopping robot, whose objective is similar to that of the HalfCheetah. The agent needs to learn to make the hopper move forward while maintaining balance and stability. The Hopper environment presents challenges related to balancing and controlling the hopping motion.
    \item[3] \textbf{Walker2d}: Walker2d is an environment controlling a two-legged robot, which resembles a simplified human walker. The goal of Walker2d is to move the walker forward while maintaining stability. walker2d poses challenges similar to HalfCheetah environment but introduces additional complexities related to humanoid structure. The above three environments have three different levels of datasets, including \texttt{medium-expert, medium-replay, medium}.
    \item[4] \textbf{AntMaze}: In the AntMaze environment, the agent controls an ant-like robot to navigate through maze-like environments to reach a goal location. The agent receives a sparse reward that the agent only receives a positive reward when it successfully reaches the goal. this makes the task more difficult. The maze configurations vary from the following environments that possess different level of complexity, featuring dead ends and obstacles. There are totally six different levels of datasets, including: \texttt{maze2d-umaze, maze2d-umaze-diverse, maze2d-medium-play, maze2d-medium-diverse, maze2d-large-play, maze2d-large-diverse}.
     \item[5] \textbf{Adroit}: In the Adroit environment, the agent controls a robotic hand to finish various manipulation tasks, including pen balancing, door opening and object relocation. The agent only receives a positive reward when the task is successfully completed, otherwise, the reward is zero, which makes the tasks more challenging. We focus on three specific tasks: the \textbf{pen} agent must manipulate a pen to keep it balanced in some orientation; the \textbf{door} agent must grasp and open a door handle; the \textbf{relocate} agent must pick up an object and move it to a target location. We adopt the binary setting of Cal\_QL~\cite{nakamoto2023calql}, which combines the cloned-level and human-level dataset. Therefore, there are tree different tasks: \texttt{pen-binary, door-binary, relocate-binary}.

\end{enumerate}

\subsection{\methodname\ performance}
\label{Apx:conter_intuition_reason}
As illustrated in Section \ref{sec:empirical_results}, \methodname\ performs best when integrated with AWAC compared to other algorithms.

The reason why AWAC-\methodname\ performs the best is detailed below. AWAC stands for advantage weighted actor critic, which is an algorithm to optimize the advantage function $A^{\pi_k}(s,a)$, while constraining the policy to stay close to offline data. AWAC does not contain any other tricks to under-estimate the value function as other offline RL algorithms~\cite{kumar2020conservative,nakamoto2023calql}, therefore AWAC could produce an accurate estimation of the values of offline data and serves as a perfect partner of \methodname. 

For the other algorithms, they adopt various techniques to achieve conservative estimation of $Q$ values in order to counteract the potential negative effects of OOD samples. Therefore, the offline guidance they provide is a little less accurate. However, these algorithms are more robust due to conservative settings and can cope with more complex tasks, as illustrated in Section 5 of CQL~\cite{kumar2020conservative}. However, it is always impossible to produce ideal $Q$ values for offline RL algorithms due to the limitations of offline datasets. The offline models trained by these algorithms could still provide guidance for the online fine-tuning process because the error of the estimation is trivial and the guidance is valuable and reliable. Furthermore, to resist the negative impact of conservative estimation, we cut the offline guidance and revert to the vanilla algorithms after a specific period of time in practice.

Overall, \methodname\ is a novel and effective paradigm, which is coherently conformed by theoretical analysis and abundant experiments.

\subsection{Cumulative regret}
\label{APx:cumulative_regret}
The cumulative regrets of the Antmaze environment of four vanilla algorithms and \methodname\ are shown in Table \ref{tab:regret}. 

It can be concluded from the table that \methodname\ possesses significantly lower regret than the vanilla algorithms, at least 40.12\% of the vanilla algorithms in the scale. This illustrate the effectiveness of our algorithms in utilizing online samples and experimentally demonstrates the superiority of \methodname\ paradigm. 

{
\setlength\fboxsep{2pt}
\begin{table*}[!t]
\centering
\begin{threeparttable}
  \small
  \caption{\textbf{Cumulative regret of online fine-tuning algorithms.}
The cumulative regret of standard base algorithms (including CQL, IQL, AWAC and Cal\_QL, denoted as ``Vanilla") compared to \methodname\ integrated algorithms (referred to as ``Ours"). The result is the average normalized score of 5 random seeds $\pm$ (standard deviation). All algorithms are conducted for 500k iterations. 
  }
  \label{tab:regret}
  \centering
  \renewcommand{\tabcolsep}{4.2pt} 
  \begin{tabular}{l|cccccccc} 
    \specialrule{0.12em}{0pt}{0pt}
	\multirow{2}{*}{Dataset}
	& \multicolumn{2}{c}{CQL} 
	& \multicolumn{2}{c}{AWAC} 
	& \multicolumn{2}{c}{IQL} 
	& \multicolumn{2}{c}{Cal\_QL}
	\\ 
	\cline{2-9}
	& Vanilla & Ours
	& Vanilla & Ours
	& Vanilla & Ours
         & Vanilla & Ours
	\\ \hline
    antmaze-u
     & 0.051\tiny{(0.005)} & {0.021}\tiny{(0.002)} & 0.081\tiny{(0.046)} & {0.080}\tiny{(0.021)} & 0.072\tiny{(0.005)} & {0.063}\tiny{(2.0)} & {0.023}\tiny{(0.003)} & 0.031\tiny{(0.002)}
    \\ 
    antmaze-ud 
    & {0.185}\tiny{(0.061)} & 0.191\tiny{(0.075)} & 0.875\tiny{(0.046)} & {0.378}\tiny{(0.090)} & 0.392\tiny{(0.116)} & {0.182}\tiny{(0.021)} & 0.142\tiny{(0.124)} & {0.133}\tiny{(0.091)}
    \\ 
    antmaze-md
    & 0.148\tiny{(0.004)} & {0.131}\tiny{(0.010)} & 1.0\tiny{(0.0)} & 1.0\tiny{(0.0)} & 0.108\tiny{(0.007)} & {0.102}\tiny{(0.008)} & {0.069}\tiny{(0.012)} & 0.076\tiny{(0.025)}
    \\ 
    antmaze-mp
    & 0.136\tiny{(0.023)} & {0.078}\tiny{(0.369)} & 1.0\tiny{(0.0)} & 1.0\tiny{(0.0)} & 0.115\tiny{(0.009)} & {0.143}\tiny{(0.020)} & {0.057}\tiny{(0.009)} & 0.071\tiny{(0.008)}
    \\ 
    antmaze-ld
    & {0.359}\tiny{(0.036)} & 0.382\tiny{(0.023)} & 1.0\tiny{(0.0)} & 1.0\tiny{(0.0)} & 0.367\tiny{(0.033)} & {0.305}\tiny{(0.041)} & {0.223}\tiny{(0.111)} &{0.219}\tiny{(0.157)}
    \\ 
    antmaze-lp
     & 0.344\tiny{(0.023)} & {0.317}\tiny{(0.052)} & 1.0\tiny{(0.0)} & 1.0\tiny{(0.0)} & 0.335\tiny{(0.032)} & {0.321}\tiny{(0.043)} & {0.203}\tiny{(0.095)} & 0.211\tiny{(0.114)}
    \\ 
    \specialrule{0.12em}{0pt}{0pt}
  \end{tabular}
\end{threeparttable}
\end{table*}
}

\subsection{Unsatisfactory performance on particular environment}
\label{Apx:unsatisfactory_performance}
We think the hyperparameter $\tau$ and the environment property may account for the unsatisfactory performance of IQL-SAMG on the environment Halfcheetah-medium-expert, rather than the integration of IQL and SAMG. 

In detail, as stated in the paper on IQL, the estimated function will gain on the optimal value as $\tau \to 1$. However, $\tau$ is not chosen to be 1 in practice and is quite low in the poorly performing environment. Additionally, this environment is relatively narrow and the training score is abnormally higher than the evaluation score. Therefore, we believe the unsatisfactory performance in this environment is just an exception and does not indicate problems of the SAMG paradigm.

\subsection{Further comparisons with hybrid RL algorithms}
\label{apx:compare_hybrid_rl}
We further compare the performance of Cal\_QL with and without \methodname\ integrated. The results are illustrated in Table{}.

\subsection{Method to get the data coverage rate of offline dataset}
\label{apx:coverage_rate}
To get the data coverage of a specific dataset, we aggregate all levels of datasets of a given environment, i.e., expert, medium-expert, medium-replay, medium, random level of datasets of environments HalfCheetah, Hopper, Walker2d. Thinking that the state and action are high-dimensional, we first perform dimensionality reduction. We uniformly and randomly select part of the data due to its huge scale and then perform t-SNE~\cite{van2008visualizing} separately on the actions and states of this subset for dimensionality reduction. Given that it is hard to model the distribution of the continuous dimensional-reduced data, We then conduct hierarchical clustering~\cite{mullner2011modern} to calculate and analyze the distribution of the data. We compute the clustering results of each environment and calculate the coverage rate based on the clustering results. To be specific, we select 10 percent of all data each time to cluster and repeat this process for 10 random seeds. For each clustering result, we calculate the data coverage rate of each level of offline dataset by counting the proportion of clustering center points. We consider one level of offline dataset to possess a clustering center if there exist more than 50 samples labeled with this clustering center. 

\section{Compute resources}
\label{apx:compute-resource}
All the experiments in this paper are conducted on a Linux server with Intel(R) Xeon(R) Gold 6226R CPU @ 2.90GHz and NVIDIA Geforce RTX 3090. We totally use 8 GPU in the experiments and each experiment takes one GPU and roughly occupies around 30\% of the GPU. It takes approximately 3 hours to 24 hours to run an experiment on one random seed, depending on the specific algorithms and environments. Specifically, the average time cost of experiments on Mujoco environments (HalfCheetah, Hopper, Walker2d) is 4.5 hours while it takes an average time of 20 hours in environment AntMaze. All experiments took a total of two months. Approximately ten days were spent on exploration, while twenty days were dedicated to completing preliminary offline algorithms.





\end{document}